\documentclass{article}

\usepackage{microtype}
\usepackage{graphicx}
\usepackage{subfigure}
\usepackage{booktabs} % for professional tables

\usepackage{hyperref}

\usepackage[accepted]{icml2025}

\usepackage{amsmath}
\usepackage{amssymb}
\usepackage{mathtools}
\usepackage{amsthm}

\usepackage[capitalize,noabbrev]{cleveref}

\theoremstyle{plain}

\theoremstyle{definition}

\theoremstyle{remark}

\usepackage[textsize=tiny]{todonotes}

\icmltitlerunning{Finding Pegasus: A Manifold-Based Approach to High-D Unsupervised Anomaly Detection}

\begin{document}

\twocolumn[
\icmltitle{Finding Pegasus: Enhancing Unsupervised Anomaly Detection in High-Dimensional Data using a Manifold-Based Approach}



\icmlsetsymbol{equal}{*}

\begin{icmlauthorlist}
\icmlauthor{R.P. Nathan}{yyy}
\icmlauthor{Nikolaos Nikolaou}{yyy}
\icmlauthor{Ofer Lahav}{yyy}

\end{icmlauthorlist}

\icmlaffiliation{yyy}{Centre for Doctoral Training in Data Intensive Science, University College London, Gower St, London WC1E 6BT, UK}
%\icmlaffiliation{comp}{Company Name, Location, Country}
%\icmlaffiliation{sch}{School of ZZZ, Institute of WWW, Location, Country}

\icmlcorrespondingauthor{R.P. Nathan}{ucaprpn@ucl.ac.uk}
%\icmlcorrespondingauthor{Firstname2 Lastname2}{first2.last2@www.uk}

\icmlkeywords{Machine Learning, ICML}

\vskip 0.3in
]



\printAffiliationsAndNotice{}  % leave blank if no need to mention equal contribution

\begin{abstract}
%The high dimensional representation of modern datasets combined with their large size means it can be a challenge to find the anomalies within them that are essential for discovery purposes, confirming predictions and identifying instrument failure.
Unsupervised machine learning methods are well suited to searching for anomalies at scale but can struggle with the high-dimensional representation of many modern datasets, hence dimensionality reduction (DR) is often performed first. 
In this paper we analyse unsupervised anomaly detection (AD) from the perspective of the manifold created in DR. 
We present an idealised illustration, `Finding Pegasus’, and a novel formal framework with which we 
categorise AD methods and their results into `on manifold' and `off manifold'. We define these terms and show how they differ. 
We then use this insight to develop an approach of combining AD methods which significantly boosts AD recall without sacrificing precision in situations employing high DR.
%We identify an effective ceiling on anomaly detection recall if relying purely on on-manifold methods, and show how this can be overcome.
When tested on MNIST data, our approach of combining AD methods improves recall by as much as 16\% compared with simply combining with the best standalone AD method (Isolation Forest), a result which shows great promise for its application to real-world data.
 
\end{abstract}

%%%%%%%%%%%%%%%%% BODY OF PAPER %%%%%%%%%%%%%%%%%%

\section{Introduction}
When it comes to availability of data, this is a golden age for researchers across a multitude of disciplines, offering huge potential to make new discoveries (find `unknown unknowns'), confirm existing or evolving theory, and highlight problems with data collection and processing.
The search for unusual or anomalous data instances is therefore a key research focus, but the sheer volume of data available combined with the high-dimensional (high-D) format in which they are often captured, mean it can be a challenge to find anomalous instances hidden within them.
These challenges have led to machine learning techniques being employed to look for anomalies at scale. In particular, unsupervised approaches are well suited to the task of finding outliers in a population of data without need for pre-labelling. Their flexibility, relative ease of implementation and the wide range of available models means unsupervised anomaly detection (AD) is employed across many fields of research. %By not encoding prior knowledge of the dataset into the search, the unsupervised method finds its own ways of separating classes and of drawing distinctions between groupings of similar instances. This can prove especially helpful in revealing `unknown unknowns' which by definition we cannot search for directly.
As we shall see, however,  unsupervised approaches tend to struggle with high-D data, and as a result dimensionality reduction (DR) is often performed directly or indirectly to produce a simpler representation of the data prior to analysis for anomalies. In doing this, a lower-dimensional manifold is created. In this paper we seek to reframe the AD problem in high-D data and encourage the reader to regard it from the perspective of this manifold.

Our paper is organised as follows. In Section~\ref{sec:AD} we provide a summary of AD concepts and terminology, and describe how the `Curse of Dimensionality' makes this task more onerous in high-D representations.
In Section~\ref{sec:DR} we see how DR is used to mitigate these issues and how the manifold thereby created is central to AD.
%, outline some of the key machine learning methods currently used for AD, with a focus on unsupervised methods. In particular, we introduce the concept of thinking of these methods as being 
In Section~\ref{sec:illus} we provide an idealisation of AD using DR -- `\textbf{Finding Pegasus}' -- 
to illustrate one of this paper's main concepts, that of AD methods working either \textbf{`on manifold'} or \textbf{`off manifold'}.
Section~\ref{sec:math_framework} approaches the topic more rigorously and provides a novel formal framework for AD with high-D data and shows how this understanding enables us to boost recall. 
\textbf{To the best of our knowledge, this is the first time the AD problem has been formulated in this way.}
Finally, in  Section~\ref{sec:MNIST} we summarise the results of applying these concepts to MNIST handwritten digits.
Where it is occasionally helpful to illustrate the concepts being discussed with a real-world example of high-D data, we shall use the example of astronomical spectra collected as part of a large galaxy survey.

%%%%%%%%%%%%%%%%%%%%%  END OF SECTION  %%%%%%%%%%%%%%%%%%%%
\section{Anomaly Detection (AD) in High-D}\label{sec:AD}

\subsection{AD Prerequisites}

Ruff et al. \yrcite{DBLP:journals/corr/abs-2009-11732} define an anomaly as `an observation that deviates considerably from some concept of normality', and we shall adopt that definition in this paper.
In AD literature, there is often also a distinction drawn between `outliers' and `novelties'. The former  is used for observations which are considered extreme or isolated within a training set of data. By contrast, the latter term is reserved for a new observation, for which similar instances were not present within the training set. 
In this context, outlier detection is considered an unsupervised activity, i.e. one in which the data does not need to be pre-labelled; whereas novelty detection might be considered supervised or semi-supervised. This work focuses on unsupervised outlier detection, though many of the conclusions may be more broadly applicable, and throughout this paper we shall use the terms `outlier' and `anomaly' interchangeably.

Ruff et al. \yrcite{DBLP:journals/corr/abs-2009-11732} also provide a  probabilistic definition of anomalies following similar treatments by Edgeworth \yrcite{edgeworth}, Barnett and Lewis \yrcite{barnettlewis} and others. Letting $ X \subseteq \mathbb{R}  ^{\scriptscriptstyle{D}} $ be the data space of dimensionality $D$ associated with some application, they define the `concept of normality' for this application as the  distribution $ \mathbb{P}^{+} $ on $ X $. This represents the `ground-truth' law of normal behaviour in this application. Correspondingly, an anomaly will be an observation $x \in X $ that is in a low probability region of $ \mathbb{P}^{+} $ and hence deviates significantly from the law of normality. They then define a set of anomalies, $ A$, as:
\begin{equation}
\label{eqn:anomalies}
   A = \{ x \in X | p^{+}(x) \leq \tau \}, \quad   \tau \geq 0
\end{equation}

\noindent where $p^{+}$ is the probability density function corresponding to $ \mathbb{P}^{+} $, and $\tau$ is a threshold set so that the probability of $A$ under $ \mathbb{P}^{+} $ is below a requisite level.

Where we have a `ground-truth' knowledge of the data, it
is possible to score the efficacy of one AD method over another in terms of standard metrics of recall, precision and F1 score as defined in Appendix~\ref{sec:stand_metr}.

\subsection{The Curse of Dimensionality }
\label{sec:unsupAD}

Whilst a probabilistic approach might be helpful in certain circumstances, it can be rendered impracticable 
where the data is represented in high-D. This is the case for many domains of research. In our astronomy example, galaxy surveys typically collect the spectra of target objects across thousands of small wavelength bins, e.g. those collected by the Dark Energy Spectroscopic Instrument \citep{DESI2016} are captured across $\approx$7800 bins, hence they are effectively 7800D. High-D data like this means we are more likely to be afflicted by the 
Curse of Dimensionality \cite{bellman1966dynamic} which impacts data distributions and densities. More background is given in Appendix~\ref{sec:CoD} but for our purposes it is sufficient to note that data points become more sparse in high dimension and so 
the concept of nearest neighbours starts to fail. 
Distributions (e.g. Gaussians), also behave differently in high dimension to our experience from low (one or two) dimensional representations. %E.g. for a multivariate Gaussian  of $D$ parameters, the distance from the centre at which the maximum concentration of points is found migrates away from zero and is found instead at $r = \sqrt{D-1}$ \citep{Delon}.
These factors have an impact on unsupervised AD methods which often rely on estimates of nearest neighbours or measurements of local concentration to determine outliers.

This is illustrated by Han et al \yrcite{han2022adbench} in a comprehensive benchmarking of AD methods available through {\tt PyOD} \citep[a Python library for detecting anomalous/outlying objects in multivariate data]{zhao2019pyod} and {\tt scikit-learn} \citep{scikit-learn}.
%Within the benchmarking, they tested 14 different unsupervised algorithms against 57 benchmark datasets.
%One of the key findings of the paper was: ``Through extensive experiments, we find surprisingly none of the benchmarked unsupervised algorithms is statistically better than others, emphasizing the importance of algorithm selection”. This is consistent with the `No Free Lunch Theorem' of machine learning \citep{wolpert}.
Review of their detailed results (in particular Table D4 from \citealt{han2022adbench}) showed that, in general, none of the unsupervised AD methods tested performed particularly well with the higher-D datasets in the benchmarking sample. A solution to this would seem to be to transform the dataset into a lower-dimensional (low-D) representation before searching for anomalies. This might occur implicitly, e.g. by downsampling or feature selection, or explicitly using the approaches described in Section~\ref{sec:detman}.

%When one examines their detailed findings, and of particular relevance to our work, we see that none of the unsupervised AD methods perform particularly well with high-D datasets with the possible exception of Fashion MNIST. 
%This can be seen in Fig. \ref{fig:Han2022Table} in which we have adapted Table D4 from \cite{han2022adbench} to summarise performance of unsupervised AD techniques on all datasets with more than 250 features. Note that the highest dimensionality of the test datasets was only 1555 (for Internet Ads) compared with approximately 7,800 for our running example of DESI spectra. So, whilst performance will be problem specific, we might expect the high-Dity of these spectra to present a challenge when looking for anomalies. 

%%%%%%%%%%%%%%%%%%%%%  END OF SECTION  %%%%%%%%%%%%%%%%%%%%

\section{Dimensionality Reduction (DR)}\label{sec:DR}

Meaningful data compression through DR relies on the `Manifold Hypothesis' (e.g. \citealt{GoodBengCour16}) that most real-world high-D datasets actually reside close to a (much) lower-D manifold which can well represent the bulk of the points in the dataset.
An $M$-dimensional manifold is part of $D$-dimensional space ($M<D$) that locally resembles an $M$-dimensional hyperplane, i.e. the surface resembles Euclidean space locally.
Points can be considered \textbf{well represented by the manifold} if there is only a small error, below some threshold, arising when we attempt to reconstruct the original high-D representation of the point from the manifold representation using whatever metric was implicit in the manifold's creation. 
Similarly, we can define a \textbf{good manifold} as one which well represents the bulk of the points considered normal in the dataset. By definition, a good manifold will satisfy the manifold hypothesis.

For example, in the case of extracting useful information from astronomical spectra, there is a body of research which shows one can construct a lower-dimensional manifold from the original high-D representation which well captures the physical properties of spectra i.e. that the manifold hypothesis is valid. Yip et al.  \yrcite{Yip_2004} found the first 3 PCA (see below) components could be used to separate different types of galaxies. Yip et al.   \yrcite{Yip_2004b_quasarclass} also found, however, that non-linear features, e.g. from quasars, required $\sim$50 PCA components to effectively represent them. For features such as these, non-linear methods might be preferred and could be very effective, e.g. Portillo et al.  \yrcite{Portillo_2020} used a variational autoencoder (VAE, see below) to reconstruct spectra well from the Sloan Digital Sky Survey (SDSS) with a manifold of only 6 latent parameters.

\subsection{Determining the Manifold}\label{sec:detman}
The two main approaches to determine an appropriate lower-dimensional manifold are projection and manifold learning.

Projection is a linear approach which assumes the data resides on or close to a lower-dimensional hyperplane in the original high-D space.
The most common projection method is Principal Component Analysis (PCA, e.g. \citealt{bishop2007}),
a statistical technique which linearly transforms the data into a new coordinate system defined by orthogonal axes which are chosen to explain as much of the variation in the data as possible.
Randomized search may be used to obtain the optimal number of dimensions.

Manifold Learning does not presume the shape of the manifold but instead attempts to learn it through parametric or non-parametric means. This approach will be advantageous where the manifold shape is more complex than a simple hyperplane.
Examples of manifold learning methods include Local Linear Embedding \citep{doi:10.1126/science.290.5500.2323}, t-SNE \citep{Maaten2008VisualizingDU}, Autoencoder (AE) \citep{1991AIChE..37..233K} and Variational Autoencoder (VAE) \citep{kingma2022autoencodingvariationalbayes}. More detail on these techniques is given in Appendix~\ref{sec:dimredmod}

%%%%%%%%%%%%%%%%%%%%%%%%%%%%%%%%%%%%%%%%%%%%%%%

\subsection{Model Dependence of the Manifold}

The exact manifold we construct will depend on the approach used, e.g.
PCA will produce an $M$-dimensional hyperplane whereas AEs or VAEs produce more complex $M$-dimensional surfaces in $D$-dimensional space.
%Except in very particular circumstances, the manifolds we identify with different dimensionality-reduction techniques will be different.
Since the manifold  is model-dependent, it follows that the \textbf{anomalies detected using it must also be model-dependent}.

\subsection{Using the Manifold for AD} \label{sec:manforAD}
There are essentially two
ways by which AD techniques use the low-D manifold to identify outliers within a population. (See Appendix~\ref{sec:OnvOffManiSchematic} for a schematic summary of this.)

%\subsubsection{`On-Manifold'}
The first approach is to look for points which are isolated or extreme in the low-D manifold representation of the distribution, typically because of distance from neighbouring points, or relative under-density of surrounding points.
%But we have already seen that in high-D this will be problematic. Hence we usually transform to low-D before searching for these points, i.e. we look for extreme and isolated points on the low-D manifold instead. 
It will be helpful to refer to methods working in this way as \textbf{on-manifold methods} and the anomalies which are detected using them as \textbf{on-manifold anomalies}. Examples of such methods include Gaussian Mixture Model (e.g. \citealt{geron2023}), K-Nearest Neighbours (\citealt{10.1145/342009.335437}), Local Outlier Factor (LOF, \citealt{10.1145/335191.335388}), Elliptic Envelope (EE, \citealt{rousseeuw}), One-Class Support Vector Machine (OCSVM, \citealt{scholkopf}), and Isolation Forest (IF, \citealt{4781136}). Details of these techniques are given in Appendix~\ref{sec:anomdetmod}.

%\subsubsection{`Off-Manifold'}
The second approach leverages the low-D manifold in a different way. Since the bulk of the dataset is well represented by the manifold, poorly represented points can be assumed to be outliers. We find these by looking for points with a high error between the original high-D representation and its reconstruction from the manifold representation employing whatever metric was used in the manifold's creation. We shall refer to these methods as \textbf{off-manifold methods} and the anomalies which are detected using them as \textbf{off-manifold anomalies}. Examples of such methods include reconstruction error from PCA, AE and VAE manifolds.

\subsection{On Manifold vs Off Manifold}

During compression to the lower-dimensional representation, some anomalous points might end up in the body of the low-D distribution rather than in extreme or isolated positions, hence on-manifold methods may not find them. This will be a particular concern when the features which make the observations abnormal are not well correlated to the features which remain after DR has been applied. Another way to think of that is that information essential to distinguishing an anomaly from a normal point is lost when the data representation undergoes compression.

Conversely, in the reconstruction approach, anomalous points which are somehow well-represented by the manifold but are in extreme or isolated positions on it, will not show significant reconstruction error and therefore will not be detected by an off-manifold method.

If  manifold construction is physically informed,  then we might be able to more easily ascribe physical meaning to the extreme points on it, and we might find these with an on-manifold method. \textbf{Hence, we might expect on-manifold anomalies to be extremes in our current thinking. Converesely, off-manifold anomalies might be where we would hope to find new science, i.e. unknown unknowns.}

Regarding the problem from the perspective of the manifold, we see there can be a difference between on- and off-manifold anomalies. If the objective is to detect as many anomalies as possible -- as it often is in discovery-driven fields -- then a useful approach would seem to be to \textbf{combine complementary on- and off-manifold methods thereby maximising the superset of anomalies which are detected}.  By `complementary' we mean that we would employ the same manifold for both the on- and off-manifold methods. 

Many researchers, mindful of the No Free Lunch Theorem \cite{wolpert}, use an ensemble of unsupervised AD techniques, thereby creating a superset of anomalies for subsequent analysis. By considering the problem from the perspective of the manifold created by DR, however, we should be able to make more informed and efficient choices about the AD methods we use.

%%%%%%%%%%%%%%%%%%%%%  END OF SECTION  %%%%%%%%%%%%%%%%%%%%

\section{Finding Pegasus}\label{sec:illus}

%We have so far used an understanding of the low-D manifold created in dimensionality reduction to propose an approach to maximise the number of anomalies we might detect by combining complementary on- and off-manifold methods. In Section \ref{sec:math_framework} we will approach this rigorously by using a simple formal framework we have devised. In this section, however, we will present an  illustration based in intuition of how we might approach anomaly detection when we are reducing dimensionality from 3D to 2D.

%\subsection{Idealized Horse Dataset}

\begin{figure}
    \centering
	\includegraphics[width=0.99\columnwidth]{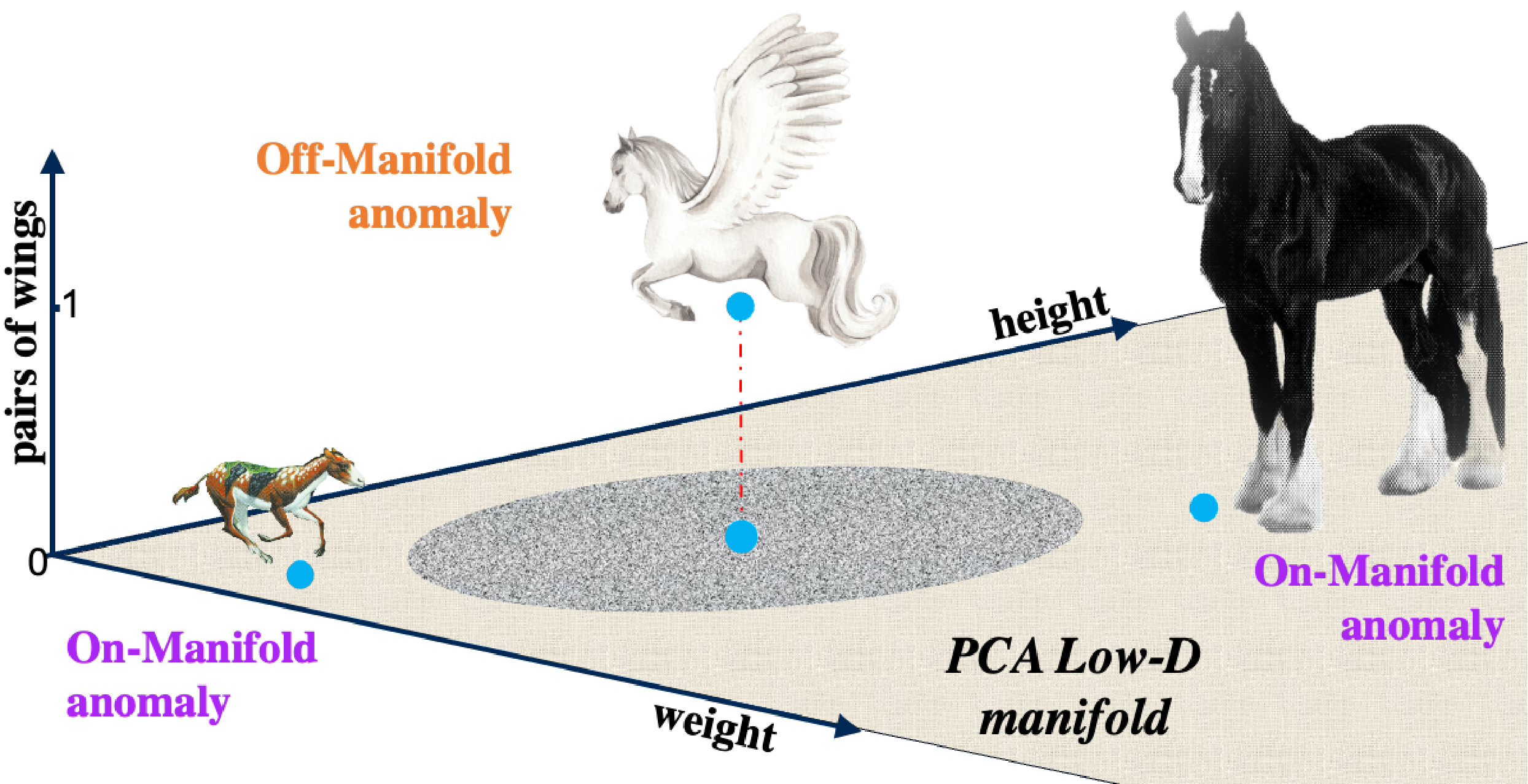}
    \caption{AD using idealised Finding Pegasus dataset illustrating the distinction between on- and off-manifold anomalies. (Horse images created using Canva.com)} 
    \label{fig:FindPeg_Conclu}
\end{figure}

We now present a simplified illustration of AD using DR. Consider an idealized horse dataset with two features, these being horse height and weight. 
From everyday experience we would anticipate good correlation between horse height and weight, so when 
plotted we might expect these data to form a distribution  like the grey shaded area in Fig.~\ref{fig:FindPeg_Conclu}.

Consider within this dataset a pair of interesting outlying points.
To the left of Fig.~\ref{fig:FindPeg_Conclu} is an artist's impression of \textit{Eohippus}, a precursor to the modern horse, thought to have been 20-30cm high. On the right is a depiction of \textit{Sampson}, thought to have been the biggest ever horse, measured at 2.2m to the shoulder. Both horses, though extreme in size, still fit in well with the overall expectation that height and weight are highly correlated.

%\subsection{Adding a Third Dimension}

We now add a third feature to the dataset, this being the number of pairs of wings the horse possesses (taking values either 0 or 1). Naturally, the existing points in the horse dataset will be unaffected by the inclusion of this extra attribute, so we add one more point to the dataset which will act as an outlier with respect to this feature. And this point, \textit{Pegasus}, the winged horse of Greek mythology, is also shown in Fig.~\ref{fig:FindPeg_Conclu}.

\subsection{DR from 3D to 2D} \label{sec:dimred3Dto2D}
Our task is to find anomalies in our dataset.
Let us assume we need to first reduce the dimensionality of the dataset down to to 2D. We choose PCA for this task. The resulting lower-dimensional manifold which is created using PCA will be a hyperplane almost coincident with the height-weight plane. %(Almost rather than exactly, since the presence of the Pegasus point will cause the manifold to lie a very small distance above the height-weight plane.)

When we compress the dataset onto this plane, %as shown in Fig.~\ref{fig:FindPeg_Mani}
most of the points, \textit{Eohippus} and \textit{Sampson} included, will be largely unchanged, since they essentially already sit on the lower-dimensional manifold ($\approx$ height-weight plane). The point representing \textit{Pegasus}, however, will be projected into the heart of the data distribution given its average height and weight and assuming its (magical) wings are weightless.
\footnote{Declaring \textit{Pegasus}'s wings to be weightless implies that this new attribute is not well correlated with existing features of the dataset, i.e. height and weight. If \textit{Pegasus}'s wings had instead been very heavy, then its total weight would have pushed its projection out of the bulk of the distribution and placed it in  a lower-density region.}

\subsection{Performing AD with the Manifold}
\label{sec:perfanomman}
%We shall now perform anomaly detection on the dataset using the low-D manifold.

%\subsubsection{`On-Manifold'}
We first perform \textbf{on-manifold} AD. 
We use Local Outlier Factor to find points which are in areas of relatively low density. 
We should find \textit{Eohippus} and \textit{Sampson} with relative ease, since they sit apart from the bulk of the data.
We are unlikely to find \textit{Pegasus}, however, because as noted above, its projection onto the manifold has placed it in the heart of a concentration of other points representing horses of average weight and height.

%\subsubsection{`Off-Manifold'}
Next we perform \textbf{off-manifold} AD, and we will look for points which are poorly represented by the low-D manifold. When we try to reconstruct the original representation of these points from their low-D manifold representation there will be a large error.
In the case of PCA, the metric used to construct the manifold is mean squared error (MSE). %i.e. the mean over all points in the dataset of the square of the Euclidean distance from the original representation of each point in 3D to its projection onto the 2D manifold ($\approx$ height-weight plane).
So, in this case, outliers will simply be those points for which the original representation is far from the 2D manifold representation.
Using this approach, we should easily find \textit{Pegasus} as an outlier.
But we are unlikely to detect \textit{Eohippus} and \textit{Sampson} as outliers because their projection onto the manifold is almost coincident with their original representation, and so will have near-zero reconstruction error.

%\subsection{Summary} \label{sec:findpegSummary}

The \textit{Eohippus} and \textit{Sampson} outliers found on manifold represent extremes in our current view of what a horse can be; whereas \textit{Pegasus}, found off manifold, is a different kind of horse entirely -- new science. All these anomalies would likely be of interest to us, and by combining complementary (i.e. using the same manifold) on- and off-manifold methods should allow us to detect all three. Henceforth, we shall refer to this as the \textbf{`Finding Pegasus Approach'}.

Had there been high correlation between wing weight and body weight 
(i.e. if \textit{Pegasus} had had heavy wings rather than magical, weightless, wings)
then we might have been able to detect \textit{Pegasus} on manifold. But because of high reconstruction error we would also have been able to detect the anomaly off manifold. Hence, in general, we do not expect the sets of anomalies found on manifold and off manifold to be disjoint.

%The second observation is that the model dependence of the anomalies will arise from 
%a) the model used to construct the low-D manifold; and b) the choice of model to look for anomalies on the manifold.  
%This model dependence ought to be mitigated, however, by choosing to combine complementary on- and off-manifold methods. We will investigate this in the next section where we will approach the problem more rigorously.

%%%%%%%%%%%%%%%%%%%%%%%%%

\begin{figure*}
    \centering
	\includegraphics[width=0.95\textwidth]{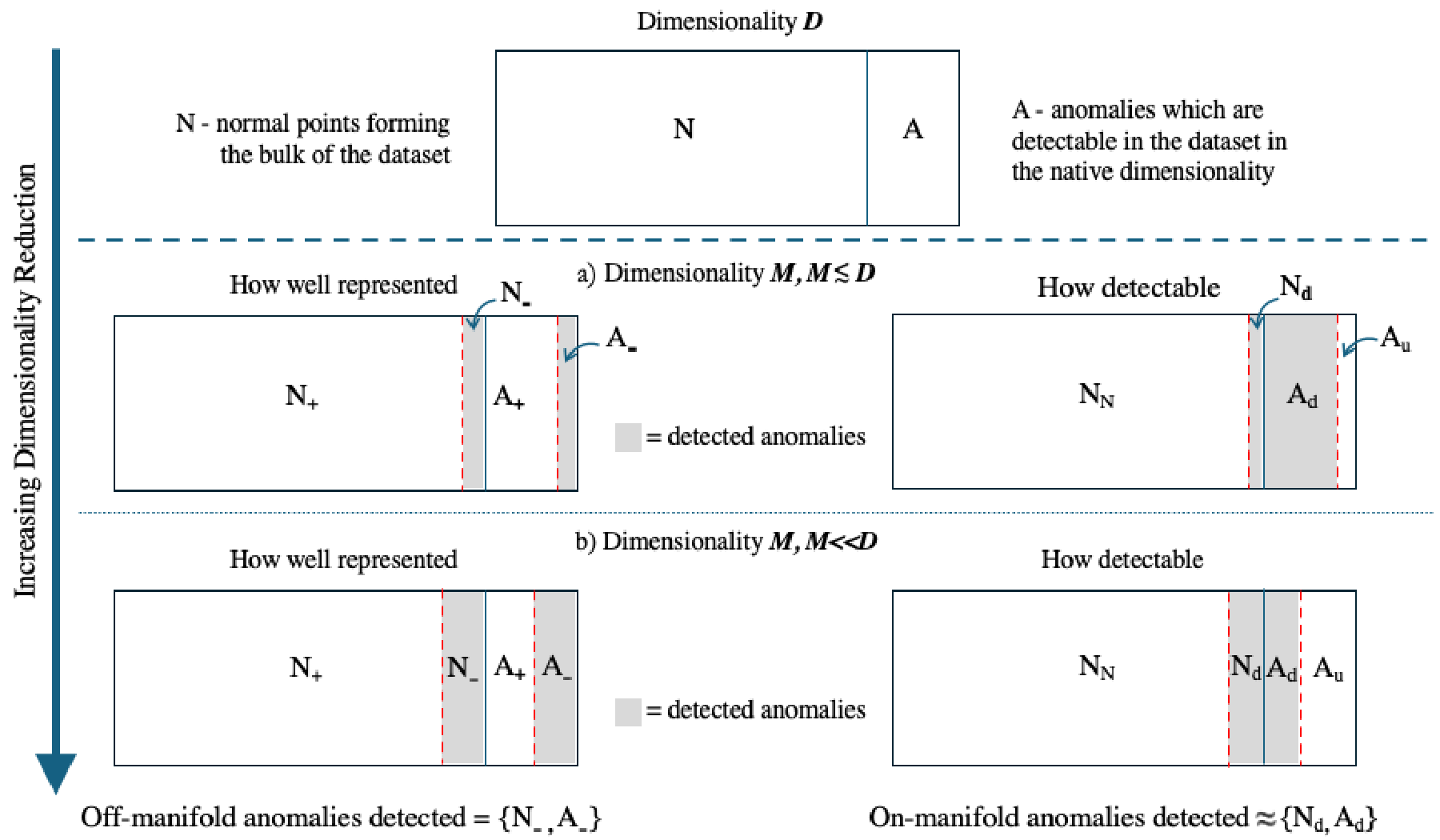}
    \caption{Schematic of formal framework for AD using DR. Detected off- and on-manifold anomalies are shown by the shaded regions.} 
    \label{fig:findpegFlowchart}
\end{figure*}

%%%%%%%%%%%%%%%%%%%%%  END OF SECTION  %%%%%%%%%%%%%%%%%%%%

\section{Formal Framework for AD in High-D Data} \label{sec:math_framework}
\subsection{Raw Data Representation}
We now introduce a simple, and we believe novel, formal framework that we have devised to describe AD in high-D data. 
%So as to keep the approach as relevant as possible to our high-D case, we have chosen not to adopt a probabilistic basis as it can be undermined by the Curse of Dimensionality.  
An overview of this framework is shown in Fig.~\ref{fig:findpegFlowchart}, and illustrative applications are provided in Appendix~\ref{sec:analysis_findpeg}.

As per Section~\ref{sec:AD} we let
$ X \subseteq \mathbb{R}  ^{\scriptscriptstyle{D}} $ be the data space of high dimensionality $D$ associated with some application.
The sample $X$ consists of $N$ \textbf{normal} points and $A$ \textbf{detectable anomalous} points, where $N$ and $A$ are disjoint, i.e. $X = N \cup A$ and $N \cap A = \emptyset$.

%\begin{equation}
%\label{eqn:D_def}
%   X = N \cup A, \qquad N \cap A = \emptyset
%\end{equation}

\noindent We regard the anomalies, $A$, as the set of points which deviate considerably from the concept of normality for our application; and conversely $N$ are the points which are in accordance with it.
By saying the points in $A$ are detectable, we mean that there is enough information contained in the raw $D$-dimensional representation of the dataset that a domain expert with requisite tools and time would be able to identify the points as being abnormal. 
Additionally, from the general definition provided by  \cite{DBLP:journals/corr/abs-2009-11732}, we can assume $|N| \gg |A|$.

%\begin{equation}
%\label{eqn:normvanom}
%   |N| \gg |A| 
%\end{equation}

\subsection{DR}
We now reduce the dimensionality of the data representation of $X$ from $D$ to $M$ dimensions, where $M<D$. In practice we would effect this by using methods of the type described in Section \ref{sec:DR}. We shall call this low-D representation $X'$.

In $X'$,  some points will be well-represented by the manifold and some will not. We shall denote points which are well represented with a subscript of `$+$'; and points which are poorly represented by `$-$'.
As per Section~\ref{sec:DR}, points will be considered well-represented by the manifold if there is only a small error arising when we attempt to reconstruct the original representation of the point from the manifold representation using whatever metric was implicit to the manifold's creation.
%Equation \ref{eqn:D_def}  now becomes:
Hence, we can write

\begin{equation}
\label{eqn:XDM_def}
   X' = (N_{+} \cup N_{-}) \cup (A_{+} \cup A_{-}),
\end{equation}

\noindent where $N_{+} \cap N_{-} =  A_{+} \cap A_{-} = \emptyset$
and for avoidance of doubt $|X'| = |X|$.
Rearranging Equation~\ref{eqn:XDM_def} we get

\begin{equation}
\label{eqn:M_red_dim}
   X' = (N_{+}  \cup A_{+}) \cup (N_{-} \cup A_{-}),
\end{equation}

\noindent where the first bracket on the right hand side contains the disjoint sets of points in $X'$ which are well-represented by the manifold, and the second bracket contains  the disjoint sets of points which are poorly represented.

\subsection{Anomalies in Reduced Dimensionality}
\label{sec:anom_RD}

\subsubsection{Detectable and Undetectable Instances}
Equation \ref{eqn:M_red_dim} indicates how \textit{well} the sets of normal and anomalous points, as labeled in the original $D$-dimensional data space, are represented by the $M$-dimensional manifold. It does not say how \textit{detectable} the anomalous points will be in lower dimension.
The reduction in dimensionality is a compression of the data space, and information must be lost as a result of this compression unless there was significant redundancy in the original representation. Hence we would expect that fewer of the anomalies will be detectable in lower dimension. If $A$ is the original set of detectable anomalies, we can write

\begin{equation}
\label{eqn:Adnu}
%   A' = A_{\scriptscriptstyle{d}} \cup  A_{\scriptscriptstyle{u}}, \qquad A_{\scriptscriptstyle{d}} \cap  A_{\scriptscriptstyle{u}} = \emptyset
   A' = A_{d} \cup  A_{u}, \qquad A_{d} \cap  A_{u} = \emptyset,
\end{equation}

\noindent where $A'$ is the representation on the $M$-dimensional manifold of the anomalous points which were detectable in $D$ dimensions, and for the avoidance of doubt $|A'| = |A|$.
$A_{d}$ are those points which are still detectable as anomalies in the lower-dimensional representation; and $A_{u}$ are those points for which there is now insufficient information to distinguish them from points considered normal. $A_{d}$ and $A_{u}$ are disjoint by definition.

The reduction in dimensionality will also sufficiently impact some normal points to make them now appear anomalous. If $N$ was the original subset of normal points, and $N'$ their representation in $M$ dimensions, we can write

\begin{equation}
\label{eqn:anom_red_dim}
%   N' =   N_{\scriptscriptstyle{N}} \cup  N_{\scriptscriptstyle{d}}, \qquad N_{\scriptscriptstyle{N}} \cap  N_{\scriptscriptstyle{d}} = \emptyset
   N' =   N_{N} \cup  N_{d}, \qquad N_{N} \cap  N_{d} = \emptyset,
\end{equation}

\noindent where for avoidance of doubt $|N'| = |N|$.
$N_{N}$ are points which as still identifiable as normal points in the low-D representation, and $N_{d}$ are points which now appear to be anomalies and would be regarded as false positives if detected through an AD process.
\footnote{ A more extensive discussion of detectability in reduced dimensionality is provided in Appendix~\ref{sec:furth_disc}.}

\subsubsection{Implications of a Good Manifold}
\label{sec:impManHyp}
We assume that efforts have been made to construct a good manifold (see Appendix~\ref{sec:furth_disc} for how this might be relaxed), hence 
the manifold hypothesis holds for our application. From Section~\ref{sec:DR}, we can assume $|N_{+}| \gg |N_{-}|$,
%\begin{equation}
%\label{eqn:norm_goodvbad}
%   |N_{+}| \gg |N_{-}| 
%\end{equation}
i.e. that most of the points we consider to be normal will be well-represented by the manifold; and that $|N_{N}| \gg |N_{d}|$,
%\begin{equation}
%   |N_{N}| \gg |N_{d}| 
%\end{equation}
i.e. that few normal points will appear anomalous in the low-D representation, regardless of the scale of DR applied (see Section~\ref{sec:DR_scale}).
It does not  follow that we can assume the same relationships hold for anomalous points since our definition of a good manifold is centred on how well it represents normal points. 
We know, however, from Equation~\ref{eqn:XDM_def} that $A' = A_{+} \cup  A_{-}$
and so combining with Equation~\ref{eqn:Adnu} we are then able to say

\begin{equation}
\label{eqn:anoms}
   A_{+} \cup  A_{-} =   A_{d}\cup  A_{u},
\end{equation}

\noindent i.e. that the union of the (disjoint) sets of well and poorly represented anomalous points will equal the union of the (disjoint) sets of anomalous points detectable on the low-D manifold and those no longer detectable. For avoidance of doubt, we note that in general $A_{+} \not \equiv   A_{d}$ and $A_{-} \not \equiv   A_{u}$.

\subsection{How Much DR?}
\label{sec:DR_scale}

We  noted in Section~\ref{sec:impManHyp} that we will be applying a level of DR such that
the manifold hypothesis holds. Hence,  we have implicitly assumed values for $M$ which allow for faithful reconstruction of the bulk of the population. But we have thus far placed no explicit constraint on the size of $M$ other than $M<D$. Going forward, it will be instructive to consider two regimes for $M$, the first where $M\lesssim D$ i.e. scenarios where M is smaller than D but of the same order of magnitude; and the second where $M\ll D$.

%\subsubsection{$M\lesssim D$}
%\label{sec:MD}
If \textbf{$M\lesssim D$} then we would straightforwardly expect $|A_{-}| \ll |A_{+}|$. The reduction in dimensionality of the dataset has been modest, hence we would expect only a minimal number of anomalous points to change from being well-represented in the native dimensionality (which all points are by definition) to being poorly represented. 
Another way of saying this is that if the level of DR is small, then $|N_{+}| \gg |N_{-}|$ holds not just because of the manifold hypothesis, but because little has changed in terms of representation. And that should be the case for anomalous points as much as for normal points. 
Following a similar argument, we would also expect $|A_{u}| \ll |A_{d}|$.
And, in the limit that $M\rightarrow D$, we would expect $|A_{-}| \rightarrow 0$ and $|A_{u}| \rightarrow 0$.

%\subsubsection{$M\ll D$}
%\label{sec:MllD}
If \textbf{$M\ll D$} then significant DR has now been performed. 
We assume the manifold hypothesis still holds, hence $|N_{+}| \gg |N_{-}|$. As noted in Section~\ref{sec:impManHyp}, it does not follow that we can write the equivalent relationship for anomalous points. It would be reasonable, however, to assume that the number of anomalies which are poorly represented by this manifold is now significant, and to write the loose constraint that $|A_{-}| \approx |A_{+}|$. 
And, following a similar argument, we would also expect $|A_{u}| \approx |A_{d}|$.

\subsection{AD in Reduced Dimensionality}

\subsubsection{On-Manifold Anomalies}\label{sec:onman_anom}
We now look for anomalies in the compressed dataset. First, we employ a model to detect anomalies on the $M$-dimensional manifold. 
This will detect $A^{*}_{mod}$ candidate anomalies, which will be a mixture of anomalies (true positives) and normal points (false positives). Hence our set of candidate on-manifold anomalies will be

\begin{equation}
\label{eqn:anoms_on}
   A^{*}_{on} =  A^{*}_{mod} = A_{mod} \cup N_{mod},
\end{equation}

\noindent where $A_{mod}$ are the true positives and $N_{mod}$  the false positives found by the model. 
We can define a \textbf{good model} to be one for which $A^{*}_{on} \approx  A_{d} \cup N_{d}$.
%\begin{equation}
%\label{eqn:anoms_ongood}
%   A^{*}_{on} \approx  A_{d} \cup N_{d}
%\end{equation}
i.e. a good model can detect most of the anomalies which are still detectable in the compressed representation on this manifold. This is not a violation of the no free lunch theorem since we require this particular model to be effective for this particular manifold representation of this particular dataset and not in general.

Since the manifold hypothesis is assumed to hold, i.e.
$|N_{N}|\gg |N_{d}|$ regardless of whether $M\lesssim D$ or $M \ll D$, but is assumed not to apply to anomalous points, it is reasonable to assert that $|N_{d}| \not \gg |A_{d}|$. As such, expert review of the set of on-manifold candidate anomalies, restored to their original dimensionality if necessary, should be feasible to identify the true anomalies. Hence, for a good model:

\begin{equation}
\label{eqn:anoms_on}
   A_{on} =  A_{mod} \approx A_{d}.
\end{equation}

\noindent We note, however, that from Section~\ref{sec:unsupAD} that we are less likely to easily meet this definition if $M \lesssim D$ since unsupervised methods can struggle with high-D data.

Since the model is detecting anomalies based on their manifold representation, it is reasonable to assume that the anomalies which
a good model would be best at finding should be similar to the set of those anomalous points which are well represented by the manifold. Hence,

\begin{equation}
\label{eqn:anoms_on_final}
   A_{on} = A_{mod} \approx A_{d} \approx   A_{+}, 
\end{equation}

\noindent i.e. using a good on-manifold model will allow us to detect on-manifold anomalies approximately equal to the set of anomalies which remain detectable in the lower-dimensional representation, and these should be similar to the set of well-represented anomalies.

\subsubsection{Off-Manifold Anomalies} \label{sec:math_offA}
Anomalies which are poorly represented by the $M$-dimensional manifold -- the off-manifold anomalies -- will be detectable by looking for points with high reconstruction error based on whatever metric was 
implicit in manifold construction. This should yield a set of candidate anomalies, 
$ A^{*}_{off}$,
which are a mixture of both normal and anomalous points, the contents of the second bracket in Equation~\ref{eqn:M_red_dim}, i.e. 

\begin{equation}
\label{eqn:anoms_off}
   A^{*}_{off} = N_{-} \cup A_{-}.
\end{equation}

\noindent We know $|N| \gg |A| $ and $|N_{+}| \gg |N_{-}|$ hence 
%From Equations~\ref{eqn:normvanom} and~\ref{eqn:norm_goodvbad} 
both $|N_{-}|$ and $|A_{-}|$ are small compared with the overall size of $X'$. Hence, expert review of the set of off-manifold candidate anomalies, restored to their original dimensionality if necessary, should be feasible to distinguish normal points from anomalous ones, thereby giving us the number of off-manifold anomalies $A_{off} =  A_{-}$, 
%\begin{equation}
%\label{eqn:anoms_off}
%A_{off} =  A_{-}
%\end{equation}
i.e. off-manifold anomalies are simply the set of anomalous points which are poorly represented by the manifold.

\subsubsection{The Finding Pegasus Approach}
Let us now combine on- and off-manifold anomalies using the two regimes of the scale of DR as described in \ref{sec:DR_scale}.
%\\[\baselineskip]

\noindent \textbf{a) $M\lesssim D$}: 
We know that as $M\rightarrow D$,  $A_{off} = A_{-} \rightarrow \emptyset$ and  $ A_{+}  \rightarrow  A'$. Thus from Equation~\ref{eqn:anoms_on_final} we see that when there is only small DR, $A_{on} \approx  A'$ for a good model and the contribution from $A_{off}$ will be small. Hence, combining off-manifold anomalies with on-manifold anomalies would not be expected to increase recall significantly. And Section~\ref{sec:unsupAD} remind us that unsupervised methods struggle in high-D, so we do not expect recall from $A_{on}$ to be high in general when raw dimensionality is high.
%\\[\baselineskip]

\noindent \textbf{b) $M\ll D$}:
From Equation \ref{eqn:anoms_on} and since $A_{off} =  A_{-}$ we see  that for a good model, the Finding Pegasus Approach would detect anomalies

\begin{equation}
\label{eqn:anoms_FindPegMeth}
   A_{det} = A_{on} \cup  A_{off} \approx
   A_{d} \cup 
    A_{-} .
\end{equation}

\noindent We know that $|A_{-}| \approx |A_{+}|$ when significant DR has been applied and that $|A_{off}| =   |A_{-}|$.
We have also already seen that $A_{on} \approx   A_{+} $. Hence $|A_{on}| \approx   |A_{off}| $ when there has been significant DR. Therefore, \textbf{relying solely on on-manifold methods will create an effective ceiling on recall where significant DR has been applied.}
In addition, \textbf{combining complementary off- and on-manifold methods should be most effective when there has been significant DR}, and we would expect them to markedly boost recall.

We have therefore demonstrated a dual benefit in using the Finding Pegasus Approach when significant DR has been applied: the contribution to recall from $A_{on}$ will be higher because the on-manifold AD method is being used in lower-D; and the contribution from $A_{off}$ will also be greater. 

\subsection{Relationship between $A_{on}$ and $A_{off}$}
\label{sec:onVoff}
Note in general  
$A_{on} \cap A_{off} \neq \emptyset $ since there might be anomalous points which are poorly represented on the low-D manifold yet are still detectable on it because of high correlation between the features which have been lost in compression and the ones that remain. This notwithstanding, \textbf{since $A_{off}$ is a consequence of the manifold construction, whilst $A_{on}$ results from the choice of AD technique used on the manifold, they are likely to represent points with very different sets of features} (as we saw in Section~\ref{sec:perfanomman}) especially when DR is significant. Thus, combining on- and off-manifold sets of anomalies as per the Finding Pegasus Approach and Equation~\ref{eqn:anoms_FindPegMeth} will not just maximize recall, but also likely increase diversity of anomalies detected.
As we have seen, however, in \ref{sec:DR_scale}, the relative size of the two sets will depend on the scale of the DR applied. For $M\lesssim D$, $|A_{off}| \ll |A_{on}|$ whereas for $M\ll D$, $|A_{off}| \approx |A_{on}|$.

%\subsection{Model Dependence in Unsupervised AD} \label{sec:mod_dep}
%As noted in Section~\ref{sec:mani}, the impact of using different models will be felt both in the creation of the lower-dimensional manifold and also in the model used to detect anomalies on it. Since we are using unsupervised methods for both tasks, the no free lunch theorem will apply, and in general, there cannot be a particular model which will work best across all datasets to generate the manifold and/or detect anomalies on it. For particular datasets, however, there may be models which work particularly well. 

%In the preceding analysis all we have assumed is a good AD model for a particular well constructed manifold with $M \ll D$, and this led us to Equation~\ref{eqn:anoms_final}. Thus overall, if we have somehow found a model which is well suited to a particular low dimensional manifold we should be able to detect most of the anomalies in our sample. Hence, across the set of well-suited model-manifold pairs we might actually be able to significantly mitigate model dependence.

%Whilst we could always construct a specific model which could act as a good model for a particular known dataset-DR instance, there is no certainty that we could find such a model easily in practice. And the general relevance of such a model would be low. Hence in practice we would resort to a model from a list of popular methods as previously discussed in this paper. 

%%%%%%%%%%%%%%%%%%%%%  END OF SECTION  %%%%%%%%%%%%%%%%%%%%

\section{Application to MNIST Handwritten Digits}\label{sec:MNIST}
To validate the concepts outlined in this paper, we applied them to an illustrative high-D dataset: MNIST  \cite{deng2012mnist}.
We chose this as we believe our approach has wide applicability, and MNIST is familiar to most researchers.
In testing, we attempted to detect images of $8$s and $7$s from a bulk population of $1$s, using on- and off-manifold methods applied to four manifolds (PCA and AE, in both $M \lesssim D$ and $M \ll D$ regimes). 
AD methods used were PCA or AE reconstruction error (RE), IF, LOF, EE and OCSVM. Methods were tested standalone, and then combined first with RE (our Finding Pegasus approach), and subsequently with IF.
Details are given in Appendix~\ref{sec:app_MNIST}, a summary of findings presented below, and the most significant results, from the $M=30$ AE manifold, are shown in Table~\ref{tab:ADtestAE_M30}.

We found that, regardless of manifold, Isolation Forest (IF) was the most effective on-manifold method for AD recall on a standalone basis.
We also observed  that for any of the four manifolds, 
combining an on-manifold AD method with the off-manifold RE method boosted recall.
This is a trivial result arising from the definition of recall, however, the mean F1 score also improved, showing that precision was not unduly sacrificed. The best performance was for the $M=30$ AE manifold where mean Finding Pegasus recall was 84\% compared to mean standalone methods of 50\%.

%%%%%%%%%%%%%%%%%%%%%%%%%%%%%%%%%%%%%%%%%%%%%%%%%%%%%%%%%%%%%%%START

\begin{table}
    \caption{AD Recall and F1 from testing MNIST with $M=30$ AE manifold. AD methods tested standalone, then combined with AE reconstruction error (RE, our Finding Pegasus approach) or IF. \newline}
    \label{tab:ADtestAE_M30}
    \centering
    \begin{tabular}{lcccccccccccc}
    %\hline
    \multicolumn{1}{l}{} &  \multicolumn{6}{c}{AE manifold, $M=30$} \\
         \cmidrule(lr){2-7}
        \multicolumn{1}{c}{\textbf{$D=784$}} & \multicolumn{2}{c}{Stand-} & \multicolumn{2}{c}{with AE RE}  & \multicolumn{2}{c}{with IF}  \\
        \multicolumn{1}{c}{} & \multicolumn{2}{c}{alone} & \multicolumn{2}{c}{\textbf{F. Pegasus}}  & \multicolumn{2}{c}{}  \\        
         \cmidrule(lr){2-3}\cmidrule(lr){4-5}\cmidrule(lr){6-7}
         AD Method &  Rec. & F1 &  Rec. & F1 & Rec. & F1 \\ 
         \hline
         \textit{Off Manifold} &  & &   &  &  &     \\ 
         \hline
         ~~AE RE & 0.73 &0.73 & n/a & n/a &  n/a & n/a \\ 
         \hline
         \textit{On Manifold} &  & &   &  &  &    \\ 
         \hline
         ~~IF  &0.60 &0.60 & \textbf{0.88} & 0.73 & n/a & n/a\\ 
         \hline
         ~~LOF &0.18 & 0.18& \textbf{0.78} & 0.56 & 0.75 & 0.51\\ 
         \hline
         ~~EE &0.59 & 0.59& \textbf{0.88} & 0.74  & 0.70 & 0.62\\ 
         \hline
         ~~OCSVM &0.41 & 0.41 & \textbf{0.83} & 0.65  & 0.61 & 0.51\\ 
         \hline
         Mean &0.50 &0.50 & \textbf{0.84} &0.67 &0.68  & 0.55\\ 
         \hline
         \hline
    \end{tabular}
\end{table}

%%%%%%%%%%%%%%%%%%%%%%%%%%%%%%%%%%%%%%%%%%%%%%%%%%%%%%%%%%%%%%%%FINISH

We also saw 
(Table~\ref{tab:ADtestAE_M30})
that with the $M=30$ AE manifold, 
\textbf{our Finding Pegasus approach of combining AD methods performed 16\% better on average than simply combining an on-manifold AD method with the best standalone on-manifold AD method, IF (84\% v 68\%)}.
%marginally better on average using PCA at high DR (76\% v 75\% at $M=84$); but 
In fact, with this manifold, all on-manifold AD methods performed significantly better using the Finding Pegasus approach than by combining with IF.

Considering the types of anomalies found, we observed that when we apply substantial DR, the proportion of $8$s detected by reconstruction error (the \textit{Pegasus} type anomalies) markedly increased, and again this was higher when using the AE manifold (88\% using AE, $M = 30$, Table~\ref{tab:det78AE}). This supports our idea that, with high DR, off-manifold methods will be effective at identifying anomalies which do not solely  reflect extremes of the bulk data (important as those might be) but which also exhibit novel features (in this case topological) where they are uncorrelated with bulk features. 

All these results demonstrate the increased effectiveness with significant DR of our Finding Pegasus approach; and also perhaps indicate that the non-linear AE creates a better manifold from the non-linear dataset than linear PCA.

Finally, we observed (e.g. Fig~\ref{fig:PCA_RL_Anomalies_v_therest}) that even when all the on-manifold methods were combined, there were some anomalies only detectable by the off-manifold method, further illustrating the importance of the Finding Pegasus approach.

\section{Summary of Key Concepts}\label{sec:concl}
%The key concepts of this paper can be summarised:

\begin{itemize}
    \item Unsupervised AD techniques can struggle with high-D data, meaning researchers often choose to work in low-D.
DR creates a manifold which is model dependent, and the anomalies detected using it are consequently also model dependent.
    \item We have reframed the unsupervised AD problem from the perspective of the lower-D manifold, encouraging thinking about unsupervised AD techniques and anomalies as being either on manifold or off manifold.
    \item Substantial DR should a) improve the efficacy of unsupervised on-manifold methods and b) increase the number of off-manifold anomalies, $|A_{off}|$.
    \item  However, the number of on-manifold anomalies detected is limited to $|A_{d}|$, so when DR is high, $|A_{on}|\approx|A_{off}|$ meaning AD relying solely on on-manifold methods will have an effective ceiling on recall.
    \item Recall will therefore be maximised by combining complementary on- and off-manifold methods (our Finding Pegasus approach) under condition of significant DR.
    \item Off-manifold anomalies are potentially of a very different character to on-manifold given they are a consequence of manifold construction rather than of the on-manifold detection technique.
\end{itemize}

We trust this paper's
illustrations and formalism will be of value to a variety of discovery-driven fields of research utilising high-D data -- e.g. astronomy, medical imaging, signal processing, financial fraud detection -- and will motivate such researchers to consider their application.

%We hope 
%the illustrations and formalism of 
%this paper will be of value to a variety of discovery-driven fields of research which make use of high-dimensional data -- e.g. astronomy, medical imaging, signal processing, financial fraud detection -- and will motivate researchers in those and other fields to consider their application.

\newpage
%%%%%%%%%%%%%%%%% END OF BODY OF PAPER which needs to fit into 8 pages %%%%%%%%%%%%%%%%%%
\section*{Acknowledgements}
We thank NOIR Lab, DESI Collaboration and Breakthrough Discuss at whose conferences early versions of this work were presented; and Chris Lintott, Constantina Nicolaou, Tara Tahseen, James Ray, Or Graur, Segev BenZvi, John Su\'arez-P\'erez, Lucy Fortson, Colin Jacobs, Antonella Palmese, Peter Melchior, Kameswara Mantha, Hayley Roberts, Koketso Mohale, Javier Via\~na P\'erez, and Bruce Bassett for insightful conversation, feedback and encouragement.

RPN has been supported by the STFC UCL Centre for Doctoral Training in Data Intensive Science (grant ST/W00674X/1) and departmental and industry contributions. OL acknowledges STFC Consolidated Grant ST/R000476/1 and visits to All Souls College and the Physics Department, University of Oxford.

%\section*{Impact Statement}

%This paper presents work whose goal is to advance the field of 
%Machine Learning. There are many potential societal consequences 
%of our work, none which we feel must be specifically highlighted here.

\nocite{langley00}

\bibliography{example_paper}
\bibliographystyle{icml2025}

%%%%%%%%%%%%%%%%%%%%%%%%%%%%%%%%%%%%%%%%%%%%%%%%%%%%%%%%%%%%%%%%%%%%%%%%%%%%%%%
%%%%%%%%%%%%%%%%%%%%%%%%%%%%%%%%%%%%%%%%%%%%%%%%%%%%%%%%%%%%%%%%%%%%%%%%%%%%%%%
% APPENDIX
%%%%%%%%%%%%%%%%%%%%%%%%%%%%%%%%%%%%%%%%%%%%%%%%%%%%%%%%%%%%%%%%%%%%%%%%%%%%%%%
%%%%%%%%%%%%%%%%%%%%%%%%%%%%%%%%%%%%%%%%%%%%%%%%%%%%%%%%%%%%%%%%%%%%%%%%%%%%%%%
\newpage
\appendix
%\onecolumn

%%%%%%%%%%%%%%%%%%%%%%%%%%%%%APPENDIX SECTION%%%%%%%%%%%%%%%%%%%%%%%%%%%%%%%%%%%%%%%%%%%%%%%%%%%%%%%%%%%%%%%%%%%%%
\section{Standard Metrics} \label{sec:stand_metr}

Where we have a good expectation of the incidence of anomalies  we are likely to find within a sample, we can
measure an AD method's effectiveness in terms of standard metrics of recall and precision as follows:

\begin{equation} \label{eqn:recall_def}
    Recall = \frac{|TP|}{|TP|+|FN|}
\end{equation}

\begin{equation} \label{eqn:prec_def}
    Precision = \frac{|TP|}{|TP|+|FP|}
\end{equation}

\noindent where $TP$ are True Positives, $FN$ are False Negatives, and $FP$ are False Positives (and $TN$ would be true negatives).

The F1 score is used as a way of combining recall and precision into a single metric and is often viewed as a measure of the all-round performance of an algorithm. It is defined from the harmonic mean of recall and precision as follows:

\begin{equation} \label{eqn:F1_def}
    F1 = 2 \times \frac{Recall \times Precision}{Recall + Precision}
\end{equation}

\section{The Curse of Dimensionality}\label{sec:CoD}

For many domains of research where the raw data representation of observations is in high-D, that representation is a consequence of how the data are collected rather than reflecting any underlying properties of the object. In our astronomy example, DESI spectra of target objects are captured across approximately 7800 small wavelength intervals called spectral channels or bins. High-D data like this is more onerous to work with computationally, and means we are likely to be afflicted by the Curse of Dimensionality.

First coined by \cite{bellman1966dynamic}, the Curse of Dimensionality refers to how our intuition into data behaviour begins to fail as we move beyond three dimensions. Two examples which are of particular importance in AD are the way that notions of \textbf{distance} and \textbf{concentration} are affected in high-D.

\subsection{Distance Metrics in high-D}
As the number of dimensions in a data representation rises, the volume of space contained in the representation also rises. As a consequence, in high-D data points become increasingly sparse and distant from one another.
E.g. if $\mathbf{x}$ and $\mathbf{y}$ are two independent variables, uniformly distributed on $[0, 1]^D$, where $D$ is the dimensionality of the space, then the mean square distance between them $\parallel \mathbf{x} - \mathbf{y} \parallel^2$ is given by \citep{Delon}:
\begin{equation}
  \mathbb{E}[\parallel \mathbf{x - y} \parallel ^2] = D/6
\end{equation}

\noindent and the standard deviation by:

\begin{equation}
   Std[\parallel \mathbf{x - y} \parallel ^2 ] \approx 0.2\sqrt{D}
\end{equation}

\noindent i.e. as the dimensionality increases, points get further apart but are relatively more tightly spread. As a result of this, 
the concept of nearest neighbours, which is fundamental to many AD techniques, starts to fail. 

\subsection{Concentration in high-D}
As well as being further away from each other as dimensionality increases, points also tend to be closer to the boundary of the space. A simple way to think of this is that in a high-D space -- i.e. one where the data is represented by many features -- it is highly likely that at least one of these features will have an extreme value. Gaussian distributions also behave differently in high dimension to our experience from low (one or two) dimensional representations. E.g. for a multivariate Gaussian  of $D$ parameters, the distance from the centre at which the maximum concentration of points is found migrates away from zero and is found instead at $r = \sqrt{D-1}$ \citep{Delon}.

%%%%%%%%%%%%%%%%%%%%%%%%%%%%%%%%%%%%%%%%%%%%%%%%%%%%%%%%%%%%%%%%%%%%%%%%%

\section{Schematic Summary of On- and Off-Manifold Anomalies when Dimensionality Reduction has been performed} \label{sec:OnvOffManiSchematic}
\begin{figure}
    \centering
	\includegraphics[width=0.99\columnwidth]{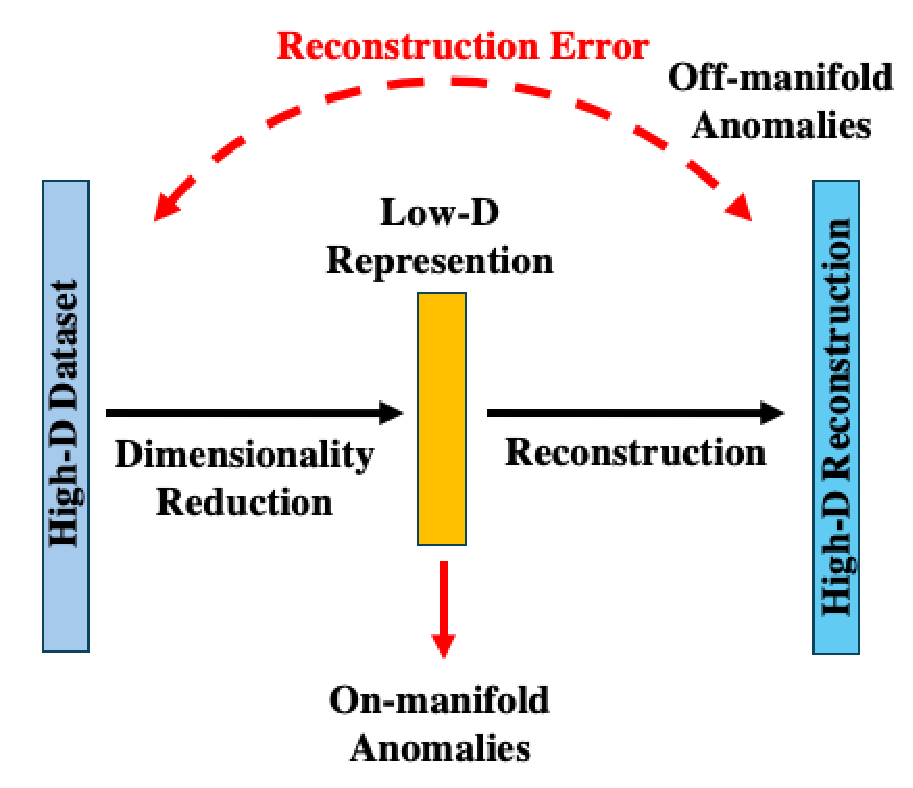}
    \caption{Schematic showing relationship between on- and off-manifold anomalies when performing AD after dimensionality reduction.}
    \label{fig:OnvOffManiSchematic}
\end{figure}

A schematic showing relationship between on- and off-manifold anomalies when performing AD after dimensionality reduction is shown in Fig.~\ref{fig:OnvOffManiSchematic}.

\section{Commonly Used Unsupervised Machine Learning Models}

\subsection{DR \label{sec:dimredmod}}

\begin{itemize}
    \item \textit{Local Linear Embedding} \citep{doi:10.1126/science.290.5500.2323}: Looks at how each training instance linearly relates to $k$ nearest neighbours and then finds a low-D representation which best preserves those local relationships.
    \item \textit{t-SNE} \citep{Maaten2008VisualizingDU}: Reduces dimensionality of the dataset while keeping similar instances close and dissimilar ones further apart. This method is widely used for the visualization in 2D or 3D of clusters existing in much higher dimensional space.
    \item \textit{Autoencoder (AE)} \citep{1991AIChE..37..233K}: A neural network which learns efficient codings of unlabelled data by compressing them into low-D. An autoencoder consists of encoder and decoder layers with a latent layer in-between. It is the size of this latent layer which sets the dimensionality to which the data is compressed. The low-D representation created by the encoder is known as the latent space. The decoder layers then attempt to reconstruct the input signal. Activation functions between layers are typically non-linear. In the special case where the activation functions are linear and there are only one encoder and decoder layers, then the autoencoder behaves like PCA. 
    \item \textit{Variational Autoencoder (VAE)} \citep{kingma2022autoencodingvariationalbayes}: A neural network similar to an autoencoder, but where the input is now mapped onto a multivariate Gaussian distribution, characterised by a set of means and standard deviations. Again, the level of compression of the original data is set by the size of the latent layer. VAEs are probabilistic in nature and are often preferred over autoencoders for their generative properties -- new points can be sampled from the latent distributions and run through the trained decoder.
\end{itemize}

\subsection{AD} \label{sec:anomdetmod}

\begin{itemize}
    \item \textit{Gaussian Mixture Model} (GMM, e.g. \citealt{geron2023}): A probabilistic model which makes the assumption that the dataset comprises points which were samples from a mixture of multiple Gaussian distributions with unknown parameters.
    \item \textit {K-Nearest Neighbours} (KNN, \citealt{10.1145/342009.335437}): Assigns an anomaly score to the data instance based on its distance to its $k$-th nearest neighbour.
    \item \textit {Local Outlier Factor} (LOF, \citealt{10.1145/335191.335388}): The algorithm compares the density of points around a given point to the density around the neighbouring points. A point is deemed an anomaly if it is more isolated (in a region of relative under-density) than its $k$ nearest neighbours. 
    \item \textit {Elliptic Envelope} (\citealt{rousseeuw}): the model assumes the data is Gaussian and learns an ellipse; outliers are outside this ellipse.
    \item \textit {One-Class Support Vector Machine} (One-Class SVM, \citealt{scholkopf}): the model attempts to learn a decision boundary which encloses the target class in feature space, thereby characterising the normal data.
    \item \textit{Isolation Forest} (IF, \citealt{4781136}): Features are randomly sub-sampled and recursively partitioned into a tree structure until all data points are isolated. Because anomalies are expected to be `few and different' they should be found closer to the root of the tree. IF is still affected by sparsity issues in high-D representations, hence a sub-space of features is often chosen, typically by ranking them by kurtosis, and these are used to construct the isolation forest. Kurtosis is sensitive to anomalies and hence a good selection criterion. But this sub-space is simply a low-D hyperplane in the original high-D space. Hence, Isolation Forest with kurtosis is simply another example of an on-manifold approach.
\end{itemize}

%%%%%%%%%%%%%%%%%%%%%%%%%%%%%%%%%%%%%%%%%%%%%%%%%%%%%%%%%%%%%%%%%%%%%%%

\section{Formal Framework for AD in High-D data -- Further Considerations} \label{sec:furth_disc}

A schematic overview of the framework described in Section~\ref{sec:math_framework} is provided in Fig.~\ref{fig:findpegFlowchart}. Further points of consideration arising from our framework are provided in the sections below.

\subsection{Undetectable Anomalies in Raw Data Representation}

We note that we could further decompose the set of normal points, $N$, as follows:

\begin{equation}
\label{eqn:N_anom}
   N = N^{N} \cup N^{A}, \qquad N^{N} \cap N^{A} = \emptyset
\end{equation}

\noindent i.e. of the points we would observe as being normal, some, $N^{N}$, are indeed normal, whilst some, $N^{A}$, are in fact anomalies, but  are undetectable even in the raw $D$-dimensional representation. The features which  make these points diverge from normality, along with any highly correlated proxies, are not contained within the raw representation, and hence it simply would not be possible to detect them, irrespective of time, tools, or expertise. Since we cannot distinguish these points using the full information at our disposal, we have no choice but to treat these points as normal.

\subsection{Detectability in Reduced Dimensionality}
\label{sec:lowDdet}

In the native dimensionality, $D$, it is natural to think of detectability of anomalies in terms of an expert manually 
sifting through an entire dataset, taking as long as needed to painstakingly inspect each instance for oddities  which would indicate it to be anomalous, They  would use visual inspection and any other tool they wished to perform this task. (The `painstaking' aspect of this process is, of course, one of the reasons why we choose to use machine learning for AD in the first place, since manual approaches are not inherently scalable.)
If an instance is identifiable by  an expert using such methods then we consider it a detectable anomaly, and if not then it is an undetectable anomaly. %We have already noted in Equation~\ref{eqn:N_anom} that an undetectable anomaly in the native dimensionality must for all practical purposes be regarded as a normal point.

%What  does detectability mean, however, in the context of the reduced dimensionality representation? Where we have performed a straightforward linear reduction in dimensionality -- such as reducing the resolution of an image or as per the illustrative Finding Pegasus example of Section~\ref{sec:illus} where we have effectively culled features -- we might reasonably expect that the aforementioned experts might be able to leverage existing methods and experience to detect anomalies in the low-D representation.

Especially where non-linear methods have been used to create the low-D manifold, however, an individual who was expert in interpreting data instances in the native dimensionality is unlikely to be familiar with their representation in lower dimension. 
%Hence, the definition of detectability used in the native dimension or when using a simple DR technique may not be easily applied. 
We therefore extend our definition of detectability to reduced dimensionality by imagining a hypothetical expert who has adequate time to become fully acquainted with the low-D representation and is able to develop and apply methods and tools which could be used to identify anomalies within it. We note that some of these might well be the very machine learning methods described in Section~\ref{sec:manforAD}.

Accordingly, we can define an anomalous point as being detectable in the low-D representation if our hypothetical expert, with access to sufficient time and tools, would be able to find it. Similarly, where we have normal points which are labeled anomalous in the low-D representation, it is because our hypothetical expert would identify them as such.

\subsection{Relaxing the Requirement for a Good AD Model for a Particular Manifold}
\label{sec:relax_good}
 Reliance on a good AD method will be mitigated where DR has been significant, since this will increase the balance of total anomalies towards off-manifold anomalies, and it should also make the on-manifold unsupervised AD methods more effective. This points to a way of further optimising the anomalies detected.

Consider that for a given manifold of dimension $M$ we use two different models, $mod_{1}$ and $mod_{2}$, then these will detect sets of anomalies $A_{mod_{1}} = A_{on_{1}}$ and $A_{mod_{2}} = A_{on_{2}}$. In general, $A_{mod_{1}}\cap   A_{mod_{2}} \neq \emptyset$
and if we have not chosen the models to be `good' models as described in Section~\ref{sec:onman_anom}, then neither $A_{on_{1}}$ nor $A_{on_{2}}$ are required or expected to be approximately equal to to $A_{d}$.

The off-manifold points are, as noted previously, a consequence of manifold construction and so will be the same in both cases.

We can extend this to use a set of models working on the $M$-dimensional manifold, $\mathbb{S} = \{ model_{1}, model_{2} ~... ~ model_{n}\} $ to form the superset of on-manifold anomalies:

\begin{equation}
\label{eqn:anoms_on_multimod}
   A_{on, \mathbb{S}} =   A_{on, 1} \cup
   A_{on, 2} \cup ... \cup A_{on, n}
\end{equation}

\noindent If we pick models which use different methodologies to detect outliers then we might expect that even for small $n$ we might be able to say $A_{on, \mathbb{S}} \approx A_{d}$, 
i.e. the set of models will perform as well as a single `good' model.

Although it might seem obvious to suggest combining on-manifold methods to boost recall, the key point here is that there is no requirement to first search for `good' models, and that especially where  $M\ll D$, and D is high, then there may not be easy to find a single model which meets our definition of `good'. Rather, we would suggest using combinations of models that employ different algorithms.

Note, however, that even with multiple models, our maximum recall will be limited to $A_{d}$ which will be problematic where there has been significant DR. In order to boost recall when $M \ll D$ further we must still use the Finding Pegasus approach so that we also include anomalous points which are poorly represented in the low-D representation.

\subsection{The Importance of the Choice of Manifold Model}
In contrast to the previous section, we should note it is still important to build a good manifold, 
 i.e. one for which the majority of points in the dataset are  well represented, thereby satisfying the manifold hypothesis and in so doing, minimizing $N_{-}$ and $N_{d}$. Without this constraint, true anomalies could be swamped by false positive normal points rendering expert review of these points much more onerous and perhaps even impossible. 
In addition, a good manifold means that off-manifold anomalies are by definition instances which differ significantly different from the feature set which characterize the bulk of the population. Where these features encode physical features, it would suggest that off-manifold anomalies could encompass instances displaying new physics (as discussed in Section~\ref{sec:illus}).

\subsection{Reconstruction from Low-D}
We note that reconstruction from the reduced dimensional representation is not possible from some commonly used DR techniques,
and so for AD purposes care should be taken in the choice of method to pick one which will allow this.

\subsection{Achieving High Recall}
It also follows from Equations \ref{eqn:anoms}, \ref{eqn:anoms_on_final} and \ref{eqn:anoms_off} that $A_{off} = A_{-} \approx   A_{u}$, i.e. 
that off-manifold anomalies, detectable through reconstruction error, will be similar to the set of anomalies which are no-longer distinguishable from normal points on the $M$-dimensional manifold. Hence:

\begin{equation}
\label{eqn:anoms_final}
   A_{on} \cup A_{off} \approx 
   A_{d} \cup 
    A_{u} = 
   A' 
\end{equation}

\noindent Hence, we have shown that where we have a good model applied to a well constructed manifold incorporating a high level of DR, using the Finding Pegasus approach (i.e. combining complementary on- and off-manifold AD techniques) might boost recall sufficiently to enable us to identify a large proportion of the anomalies in $X$. 

\section{Applying the Formal Framework} \label{sec:analysis_findpeg}
We now demonstrate how the formal framework described in Section~\ref{sec:math_framework} might be used in practice. First we apply the framework  to the idealized dataset depicted in the Finding Pegasus illustration. Then we shall show how it might be used to derive generalised expressions for precision and recall.

\subsection{Analysis of Finding Pegasus Illustration}
\subsubsection{Overview}
For the purposes of this analysis, and reflecting the discussion in Section~\ref{sec:dimred3Dto2D}, we shall include \textit{two} \textit{Pegasus} points, one with magical (i.e. weightless) wings and one with heavy wings.
A summary of how these points would be represented using our framework is shown in Table~\ref{tab:peg}.

The first two columns of the table reflect how well or poorly the anomalous points are represented on the lower-dimension manifold.  The third and fourth columns indicate whether the anomalous points are detectable or undetectable on the lower-dimensional manifold.
%The dimensionalities of our illustration are made explicit with the superscript `$3,2$' of the anomaly categories  meaning three dimensions reduced to two.

We note that anomalous points which are poorly represented on the manifold are off-manifold anomalies by definition and are given by the second column in Table~\ref{tab:peg}.

Making the assumption that we can find a `good' model as described in Section~\ref{sec:onman_anom} to search for anomalies on the low-D  manifold we have created (2D in this example) then we would expect to find most of the detectable anomalies. Hence the on-manifold anomalies will be the set of points appearing in the third column.

\begin{table}
    \caption{Finding Pegasus illustration categorized according to framework in Section~\ref{sec:math_framework} \newline}
    \label{tab:peg}
    \centering
    \begin{tabular}{lccccc}
    \hline
         &  $\in A_{+}$ & $\in A_{-}$  &$|$& $\in A_{d}$ & $\in A_{u}$ \\ 
         \hline
         \textit{Eohippus}& \checkmark &  &$|$& \checkmark & \\ 
         \textit{Sampson}& \checkmark &  &$|$& \checkmark & \\ 

         \textit{Pegasus} &  &  &$|$&  & \\ 
           ~~ - magical wings&  & \checkmark &$|$&  & \checkmark\\ 
           ~~ - heavy wings&  & \checkmark &$|$& \checkmark & \\ 
         \hline
         \hline
    \end{tabular}
\end{table}

\subsubsection{\textit{Eohippus} and \textit{Sampson}}
Taking \textit{Eohippus} and \textit{Sampson} first, we see both points are  categorized as being well represented on the manifold and also being detectable on the manifold. Because of the latter, they will be part of the set of on-manifold anomalies.

\subsubsection{$Pegasus_{m}$ and $Pegasus_{h}$}
\textit{Pegasus} with magical wings, $Pegasus_{m}$, is poorly represented on the manifold, hence an off-manifold anomaly by definition. Because the wing feature had no correlation with the two retained features of height and weight it was projected down into the centre of the distribution of normal horse points. Hence $Pegasus_{m}$ is unlikely to  be detectable on the manifold.

\textit{Pegasus} with heavy wings, $Pegasus_{h}$, is also poorly represented on the manifold, hence an off-manifold anomaly by definition. Now though, its heavy wings will mean it is no longer projected into the centre of the distribution of normal horse points but will be found in a lower density region representing horses with abnormally high weight. Thus $Pegasus_{h}$ should be detectable on the manifold.

In summary then we will have:

\begin{equation}
\label{eqn:Aon}
   A_{on}  = \{ \textit{Eohippus}, \textit{Sampson}, \textit{Pegasus}_{h} \}
\end{equation}

\noindent and 

\begin{equation}
\label{eqn:Aoff}
   A_{\scriptscriptstyle{off}} =
   \{ \textit{Pegasus}_{m}, \textit{Pegasus}_{h} \}
\end{equation}

\noindent We can therefore see that \textit{Pegasus}$_{h}$ will be an element of the sets of both on- and off-manifold anomalies.
Consequently, and as discussed in Section~\ref{sec:onVoff}:

\begin{equation}
\label{eqn:AonAoffnotemptyset}
   A_{on} \cap A_{off} \neq \emptyset 
\end{equation}

%%%%%%%%%%%%%%%%%%%%%%%

\subsection{Derivation of Expressions for Recall and Precision} \label{sec:analysis_precrec}
We now apply the formal framework to derive expressions for recall and precision metrics as introduced in Section~\ref{sec:AD} and defined in equations \ref{eqn:recall_def} and \ref{eqn:prec_def}  respectively.

\subsubsection{On Manifold}
For on-manifold anomalies we assume we are using a good model with $M \ll D$ on a well constructed manifold as previously defined and so can say, knowing $A^{*}_{on} \approx  A_{d} \cup N_{d}$ that:

\begin{equation}
    \label{eqn:TPon}
   TP_{on} = A_{mod} \approx A_{d}
\end{equation}

\noindent and 

\begin{equation} \label{eqn:FP_on}
   FP_{on} = N_{mod} \approx N_{d}
\end{equation}

\noindent Since $TP_{on} \cup FN_{on} = A'$, i.e. the union of the sets of true positives and false negatives must equal the entire set of anomalies, then using Equation~\ref{eqn:Adnu} we can say the anomalous points not captured must be:

\begin{equation}
   FN_{on} \approx A_{u}
\end{equation}

Thence:

\begin{equation} \label{eqn:recall_on}
    Recall_{on} \approx \frac{|A_{d}|}{|A_{d}|+|A_{u}|} = \frac{|A_{d}|}{|A'|}
\end{equation}

\begin{equation} \label{eqn:prec_on}
    Precision_{on} \approx \frac{|A_{d}|}{|A_{d}|+|N_{d}|}
\end{equation}

\noindent From Equation \ref{eqn:recall_on} we can see that recall will be limited by $|A_{d}|$ if we are only using on-manifold methods. And from  Equation \ref{eqn:prec_on} we can see that a well-constructed manifold which minimises $N_{d}$ will maximise precision.

\subsubsection{Off manifold}
For off-manifold anomalies we can say, following the arguments in Section~\ref{sec:math_offA}, and noting that False Negatives will simply be anomalous points not captured using this method:

\begin{equation}
   \label{eqn:TPoff}
   TP_{off} = A_{-} 
\end{equation}

\begin{equation} \label{eqn:FP_off}
   FP_{off} = N_{-} 
\end{equation}

\begin{equation}
   FN_{off} = A_{+} 
\end{equation}

\noindent Hence we can say:

\begin{equation} \label{eqn:recall_off}
    Recall_{off} = \frac{|A_{-}|}{|A_{-}|+|A_{+}|} = \frac{|A_{-}|}{|A'|}
\end{equation}

\begin{equation} \label{eqn:prec_off}
    Precision_{off} = \frac{|A_{-}|}{|A_{-}|+|N_{-}|}
\end{equation}

\noindent From Equation \ref{eqn:prec_off} we can see that a well constructed manifold which captures most of the normal points and minimizes $N_{-}$  will maximize the precision of off-manifold AD.

Hence, precision for both on- and off-manifold methods are maximized by building a good manifold.

\subsubsection{Finding Pegasus approach}
We now combine on- and off-manifold methods to determine precision and recall for the Finding Pegasus approach. First let us define the associated metrics $TP_{f}$, $FP_{f}$ and $FN_{f}$. From Equations~\ref{eqn:TPon} and~\ref{eqn:TPoff} and using Equation~\ref{eqn:anoms_final} we can see:

\begin{equation}
   TP_{f} \approx 
   A_{d} \cup A_{-} \approx
    A_{d} \cup A_{u} =
   A' 
\end{equation}

\noindent It follows immediately that

\begin{equation}
   FN_{f} \approx \emptyset 
\end{equation}

\noindent And from Equations~\ref{eqn:FP_on} and \ref{eqn:FP_off}:

\begin{equation}
   FP_{f} \approx N_{d} \cup N_{-}
\end{equation}

Hence:

\begin{equation} \label{eqn:recall_combi}
    Recall_{f} \approx \frac{|A'|}{|A'|+|\emptyset|} = 1
\end{equation}

\noindent i.e. recall for this method will approach 100\% when we have a good AD method applied to a well constructed manifold as previously implied by Eqn~\ref{eqn:anoms_final}.

Similarly we can write:
\begin{equation} \label{eqn:recall_combi}
    Precision_{f} \approx \frac{|A'|}{|A'| + |N_{d} \cup N_{-}|}
\end{equation}

%as a consequence of Equations \ref{eqn:normvanom}, \ref{eqn:norm_goodvbad} and \ref{eqn:anoms_on_moreathann}:
\noindent If we have maximised precision of both on- and off-manifold anomalies, by minimising
$|N_{d}|$ and  $|N_{-}|$
then we will have also maximised precision overall. 
Hence, the Finding Pegasus approach should deliver maximum recall without undue penalty to precision.

%%%%%%%%%%%%%%%%%%%%%%%%%%%%%APPENDIX SECTION%%%%%%%%%%%%%%%%%%%%%%%%%%%%%%%%%%%%%%%%%%%%%%%%%%%%%%%%%%%%%%%%%%%%%

\begin{table*}
    \caption{Summary of AD testing on MNIST Handwritten Digits using a variety of on-manifold methods in conjunction with PCA manifolds generated at two different levels of DR, first standalone and then combined with PCA Reconstruction Error (the Finding Pegasus Approach) and  Isolation Forest methods in turn.  \newline}
    \label{tab:ADtestPCA}
    \centering
    \begin{tabular}{lcccccccccccc}
    %\hline
    \multicolumn{1}{l}{$D=784$} & \multicolumn{6}{c}{PCA manifold, $M=245$}  & \multicolumn{6}{c}{PCA manifold, $M = 84$} \\
         \cmidrule(lr){2-7}\cmidrule(lr){8-13}
        \multicolumn{1}{c}{} & \multicolumn{2}{c}{Standalone} & \multicolumn{2}{c}{with PCA RE}  & \multicolumn{2}{c}{with IF} & \multicolumn{2}{c}{Standalone} & \multicolumn{2}{c}{with PCA RE}  & \multicolumn{2}{c}{with IF} \\
        \multicolumn{1}{c}{} & \multicolumn{2}{c}{} & \multicolumn{2}{c}{\textbf{F. Pegasus}}  & \multicolumn{2}{c}{} & \multicolumn{2}{c}{} & \multicolumn{2}{c}{\textbf{F. Pegasus}}  & \multicolumn{2}{c}{} \\
         \cmidrule(lr){2-3}\cmidrule(lr){4-5}\cmidrule(lr){6-7}\cmidrule(lr){8-9}\cmidrule(lr){10-11}\cmidrule(lr){12-13}
         AD METHOD&  Rec. & F1 &  Rec. & F1 & Rec. & F1 & Rec. & F1  & Rec. & F1& Rec. & F1\\ 
         \hline
         \textit{Off Manifold} &  & &   &  &  &  & &   \\ 
         \hline
         ~~Reconstruction Error & 0.53 & 0.53&  n/a & n/a & n/a & n/a &0.65 &0.65 & n/a & n/a &  n/a & n/a \\ 
         \hline
         \textit{On Manifold} &  & &   &  &  &  & &   \\ 
         \hline
         ~~Isolation Forest & 0.63& 0.63& \textbf{0.70} &0.61 & n/a & n/a & 0.67 &0.67 & \textbf{0.77} & 0.70 & n/a & n/a\\ 
         \hline
         ~~LOF &0.29 &0.29 &  \textbf{0.71} & 0.52 & 0.76 & 0.56 &0.25 & 0.25& \textbf{0.77} & 0.56 & 0.79 & 0.57\\ 
         \hline
         ~~Ellipticenvelope &0.58 & 0.58&  \textbf{0.72} & 0.57 & 0.77 & 0.62 &0.37 & 0.37& \textbf{0.75} & 0.56  & 0.73 & 0.55\\ 
         \hline
         ~~One-Class SVM &0.60 &0.59 &  \textbf{0.72} & 0.60 & 0.73 & 0.63&0.56 & 0.55 & \textbf{0.77} & 0.65  & 0.73 & 0.63\\ 
         \hline
         Mean Recall/F1 &0.53 & 0.52&  \textbf{0.71} & 0.58 & 0.75 & 0.61 &0.50 &0.50 & \textbf{0.76} &0.62 &0.75  & 0.59\\ 
         \hline
         \hline
    \end{tabular}
\end{table*}

\section{Application of Finding Pegasus Approach to MNIST data}
\label{sec:app_MNIST}

\subsection{Choice of Dataset}
The  MNIST \citep[modified National Institute of Standards and Technology,][]{deng2012mnist} dataset comprises 60,000 training images and 10,000 test images of handwritten digits of the numbers $0$ to $9$ in roughly equal quantities, along with corresponding sets of labels. These black and white digits are size normalized to 28x28 pixels giving the dimensionality, $D$, of each flattened image vector as 784, high enough to be able to test the Finding Pegasus Approach. Whilst MNIST might often be regarded as an introductory dataset for supervised classification purposes, it still contains much subtlety which captures well the issues being faced in unsupervised AD in high-D data. 

For the purposes of this study we worked with a subset of the test data. We chose the number $1$s (comprising approximately 6,700 data points) as the bulk data (the normal points) and contaminated it with anomalies consisting 120 points ($\approx 2$\%) of the $7$s dataset to represent an extreme version of the bulk topology; and 120 points ($\approx$2\%) of the $8$s dataset to introduce a class of anomalies with a very different topology, in this case consisting of two conjoined loops. In the parlance of this paper, the $7$s might be considered anomalies of the \textit{Eohippus} or \textit{Sampson} type; whereas the $8$s could be regarded as anomalies of the \textit{Pegasus} type. 
Examples of the MNIST handwritten digits used in this work are shown below in Figs.~\ref{fig:Ones},~\ref{fig:Sevens} and~\ref{fig:Eights}.  

\begin{figure}
    \centering
	\includegraphics[width=0.99\columnwidth]{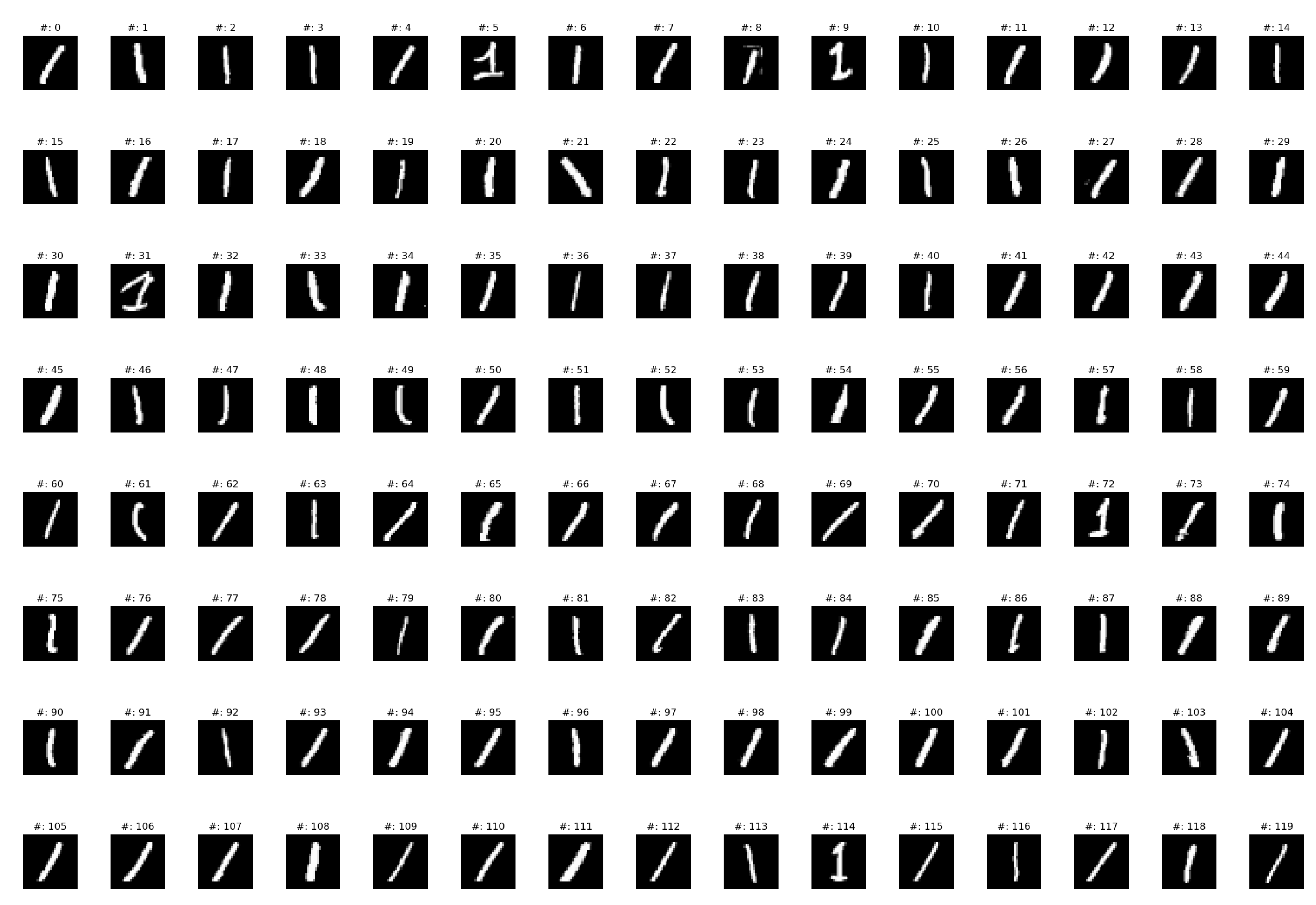}
    \caption{Sample of $1$s from MNIST handwritten digits dataset representing `normal' points.     
    }
    \label{fig:Ones}
\end{figure}

\begin{figure}
    \centering
	\includegraphics[width=0.99\columnwidth]{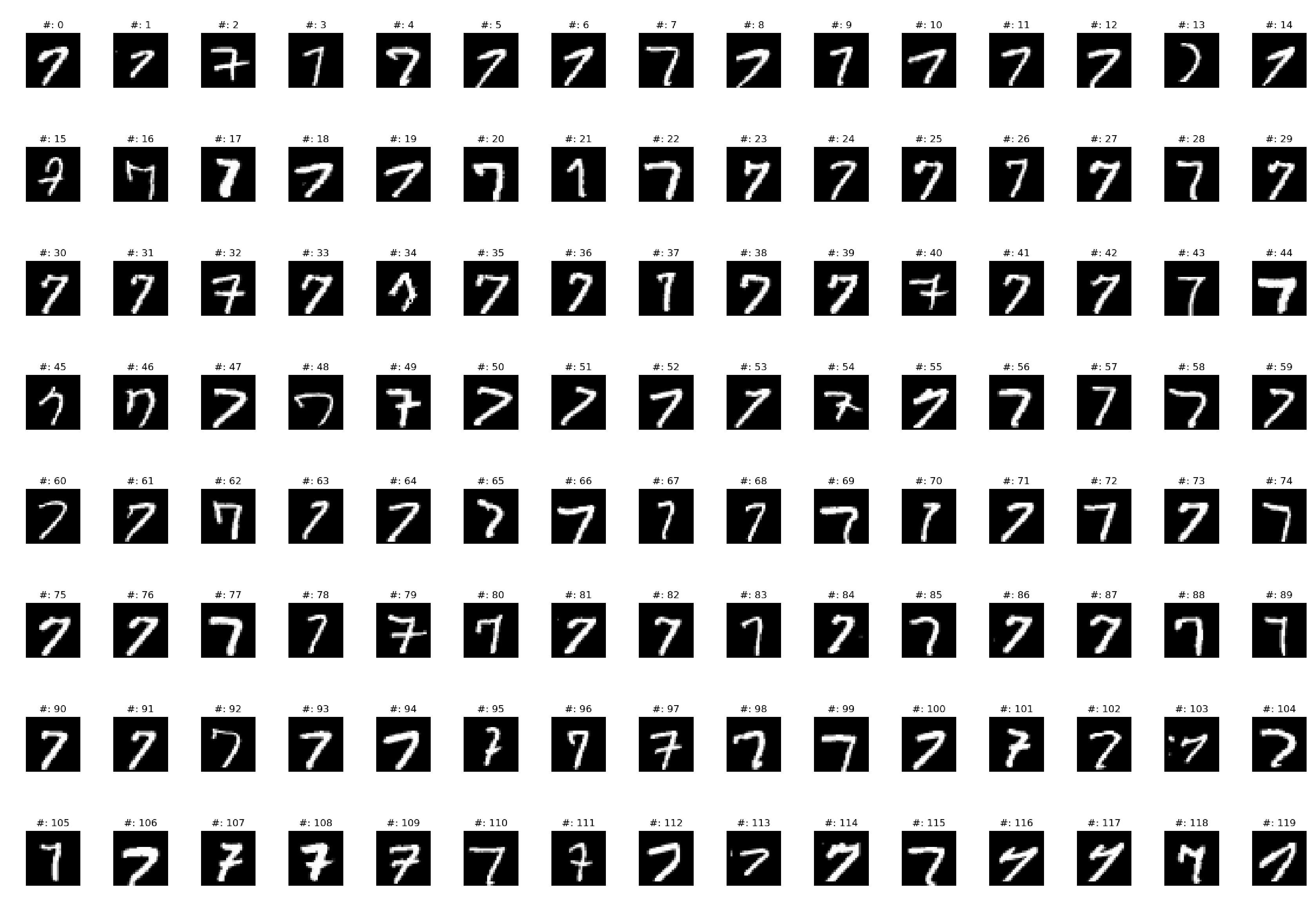}
    \caption{$7$s `anomalies' from MNIST handwritten digits dataset representing extreme forms of the bulk topology - \textit{Eohippus} and \textit{Sampson} points.   
    }
    \label{fig:Sevens}
\end{figure}

\begin{figure}
    \centering
	\includegraphics[width=0.99\columnwidth]{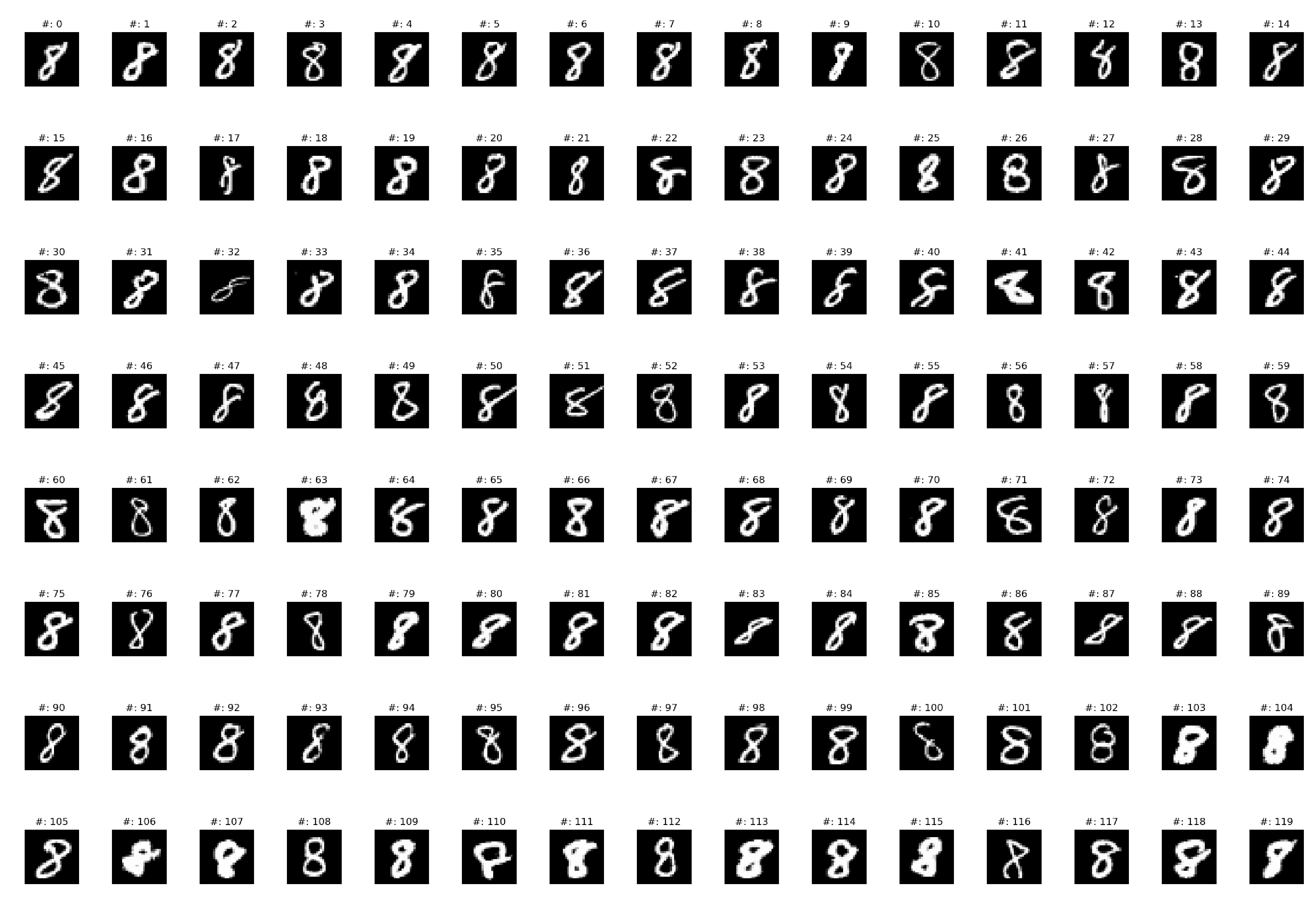}
    \caption{$8$s `anomalies' from MNIST handwritten digits dataset representing a radically different topology - \textit{Pegasus} points.
    }
    \label{fig:Eights}
\end{figure}

Although $7$s and $8$s are the anomalies we are principally trying to detect, it is worth noting that even within the $1$s dataset there are a minority which deviate from the basic form of a single thin vertical line. In particular, there are more complex `peaked' forms (made up of two thin lines: / + $|$ ) and  `peak-and-foot' forms (made up of three thin lines: / + $|$ + $-$) which in different contexts might also be viewed as anomalous. We did not attempt to quantify these instances separately here as to do so would require performing a sub-classification of the MNIST $1$s dataset, which would have been subjective and not the purpose of this paper. 

We note also that we are not seeking to present here the state of the art regarding AD within this dataset, but merely to demonstrate with data how the methods discussed in this paper might be beneficial in practice.

\begin{table*}
    \caption{Summary of AD testing on MNIST Handwritten Digits using a variety of on-manifold methods in conjunction with AE manifolds generated at two different levels of DR, first standalone and then combined with AE Reconstruction Error (the Finding Pegasus Approach) and  Isolation Forest methods in turn.  \newline}
    \label{tab:ADtestAE}
    \centering
    \begin{tabular}{lcccccccccccc}
    %\hline
    \multicolumn{1}{l}{$D=784$} & \multicolumn{6}{c}{AE manifold, $M=245$}  & \multicolumn{6}{c}{AE manifold, $M = 30$} \\
         \cmidrule(lr){2-7}\cmidrule(lr){8-13}
        \multicolumn{1}{c}{} & \multicolumn{2}{c}{Standalone} & \multicolumn{2}{c}{with AE RE}  & \multicolumn{2}{c}{with IF} & \multicolumn{2}{c}{Standalone} & \multicolumn{2}{c}{with AE RE}  & \multicolumn{2}{c}{with IF} \\
        \multicolumn{1}{c}{} & \multicolumn{2}{c}{} & \multicolumn{2}{c}{\textbf{F. Pegasus}}  & \multicolumn{2}{c}{} & \multicolumn{2}{c}{} & \multicolumn{2}{c}{\textbf{F. Pegasus}}  & \multicolumn{2}{c}{} \\        
         \cmidrule(lr){2-3}\cmidrule(lr){4-5}\cmidrule(lr){6-7}\cmidrule(lr){8-9}\cmidrule(lr){10-11}\cmidrule(lr){12-13}
          AD METHOD &  Rec. & F1 &  Rec. & F1 & Rec. & F1 & Rec. & F1  & Rec. & F1& Rec. & F1\\ 
         \hline
         \textit{Off Manifold} &  & &   &  &  &  & &   \\ 
         \hline
         ~~Reconstruction Error & 0.45 & 0.45&  n/a & n/a & n/a & n/a &0.73 &0.73 & n/a & n/a &  n/a & n/a \\ 
         \hline
         \textit{On Manifold} &  & &   &  &  &  & &   \\ 
         \hline
         ~~Isolation Forest & 0.67& 0.67& \textbf{0.71} &0.58 & n/a & n/a &0.60 &0.60 & \textbf{0.88} & 0.73 & n/a & n/a\\ 
         \hline
         ~~LOF &0.41 &0.41 &  \textbf{0.70} & 0.50 & 0.85 & 0.64 &0.18 & 0.18& \textbf{0.78} & 0.56 & 0.75 & 0.51\\ 
         \hline
         ~~Ellipticenvelope &0.45 & 0.45&  \textbf{0.73} & 0.55 & 0.86 & 0.66 &0.59 & 0.59& \textbf{0.88} & 0.74  & 0.70 & 0.62\\ 
         \hline
         ~~One-Class SVM &0.60 &0.60 &  \textbf{0.66} & 0.53 & 0.75 & 0.67&0.41 & 0.41 & \textbf{0.83} & 0.65  & 0.61 & 0.51\\ 
         \hline
         Mean Recall/F1 &0.50 & 0.52&  \textbf{0.76} & 0.54 & 0.75 & 0.66 &0.50 &0.50 & \textbf{0.84} &0.67 &0.68  & 0.55\\ 
         \hline
         \hline
    \end{tabular}
\end{table*}

\begin{table*}
    \caption{Summary of AD testing on MNIST Handwritten Digits showing the number of off-manifold and on-manifold anomalies detected at two different levels of DR using PCA manifolds. $7$s can be considered \textit{Eohippus}-type anomalies.  \newline}
    \label{tab:det78PCA}
    \centering
    \begin{tabular}{lcccccccccccc}
    %\hline
    \multicolumn{1}{l}{$D=784$} & \multicolumn{4}{c}{PCA manifold, $M=245$}  & \multicolumn{4}{c}{PCA manifold, $M = 84$} \\
         \cmidrule(lr){2-5}\cmidrule(lr){6-9}
         
         \multicolumn{1}{c}{} & \multicolumn{2}{c}{$7$s} & \multicolumn{2}{c}{$8$s}  & \multicolumn{2}{c}{$7$s} & \multicolumn{2}{c}{$8$s}  \\
         \cmidrule(lr){2-3}\cmidrule(lr){4-5}\cmidrule(lr){6-7}\cmidrule(lr){8-9}
         
          AD METHOD &  detected & Recall &  detected & Recall & detected & Recall & detected & Recall  \\ 
         \hline
         \textit{Off Manifold} &  & &   &  &  &  & &   \\ 
         \hline
         ~~ Reconstruction Error & 86 & 0.72&  41 & 0.34 & 89 & 0.74 &67 &\textbf{0.56}   \\ 
         \hline
         \textit{On Manifold} &  & &   &  &  &  & &   \\ 
         \hline         
         ~~ Isolation Forest & 79& 0.66& 72 &0.60 & 71 & 0.59 &89 &0.74 \\ 
         \hline
         ~~ LOF &28 &0.23 &  42 & 0.35 & 24 & 0.20 &36 & 0.30\\ 
         \hline
         ~~ Ellipticenvelope &85 & 0.71&  55 & 0.46 & 36 & 0.30 &53 & 0.44\\ 
         \hline
         ~~ One-Class SVM &67 &0.56 &  78 & 0.65 & 62 & 0.52&72 & 0.60 \\ 
         \hline
         Mean On Manifold &65 & 0.54&  62 & 0.51 & 48 & 0.40 &63 &0.52 \\ 
         \hline
         \hline
    \end{tabular}
\end{table*}

\begin{table*}
    \caption{Summary of AD testing on MNIST Handwritten Digits showing the number of off-manifold and on-manifold anomalies detected at two different levels of DR using AE manifolds. $8$s can be considered \textit{Pegasus}-type anomalies. \newline}
    \label{tab:det78AE}
    \centering
    \begin{tabular}{lcccccccccccc}
    %\hline
    \multicolumn{1}{l}{$D=784$} & \multicolumn{4}{c}{AE manifold, $M=245$}  & \multicolumn{4}{c}{AE manifold, $M = 30$} \\
         \cmidrule(lr){2-5}\cmidrule(lr){6-9}
         \multicolumn{1}{c}{} & \multicolumn{2}{c}{$7$s} & \multicolumn{2}{c}{$8$s}  & \multicolumn{2}{c}{$7$s} & \multicolumn{2}{c}{$8$s}  \\
         \cmidrule(lr){2-3}\cmidrule(lr){4-5}\cmidrule(lr){6-7}\cmidrule(lr){8-9}      
          AD METHOD &  detected & Recall &  detected & Recall & detected & Recall & detected & Recall  \\        
         \hline
         \textit{Off Manifold} &  & &   &  &  &  & &   \\ 
         \hline
         ~~ Reconstruction Error & 34 & 0.28&  73 & 0.61 & 70 & 0.58 &105 &\textbf{0.88}   \\ 
         \hline
         \textit{On Manifold} &  & &   &  &  &  & &   \\ 
         \hline         
         ~~ Isolation Forest & 58& 0.48& 102 &0.85 & 64 & 0.53 &79 &0.66 \\ 
         \hline
         ~~ LOF &48 &0.40 &  51 & 0.43 & 24 & 0.20 &18 & 0.15\\ 
         \hline
         ~~ Ellipticenvelope &85 & 0.71&  24 & 0.20 & 72 & 0.60 &70 & 0.58\\ 
         \hline
         ~~ One-Class SVM &62 &0.52 &  82 & 0.68 & 50 & 0.42&49 & 0.41 \\ 
         \hline
         Mean On Manifold &63 & 0.53&  65 & 0.54 & 43 & 0.44 &54 &0.45 \\ 
         \hline
         \hline
    \end{tabular}
\end{table*}

\subsection{Summary of Testing Performed}
In order to test the ideas of this paper, we used both linear and non-linear methods to reduce the dimensionality of the $\approx7000$ data points comprising the dataset described above and build lower-dimensional manifolds from them. The methods used were PCA projection (linear) and AE manifold learning (non-linear). 
For each, we chose two levels of DR to roughly correspond to the $M\lesssim D$ and $M\ll D$ regimes introduced in Section~\ref{sec:DR_scale} and which satisfied the manifold hypothesis of achieving a good lower-dimensional representation of the dataset. 
More detail on manifold construction is given in Section~\ref{sec:maniconst} below. 
These resulting lower-dimensional manifolds were then used for off-manifold and on-manifold AD.

Off-manifold AD was performed by measuring the MSE between a data point's original representation and the reconstruction from its manifold representation. Anomalies were defined as the least-well represented 240 points detected, i.e. those with the highest reconstruction error.

On-manifold AD was performed using the Scikit-learn Python implementation of four different commonly used techniques: Local Outlier Factor, Isolation Forest, One-Class Support Vector Machine and EllipticEnvelope. In each case anomalies were defined as the 240 most outlying points using whichever metric was implicit to the method.

For the sets of anomalies detected using these different methods, we calculated recall, precision and F1 metrics, as previously defined, with detected $7$s and $8$s being regarded as True Positives and all detected $1$s being treated as False Positives, regardless of whether we thought them to be unusual forms of the kind mentioned above.

\begin{figure*}
    \centering
	\includegraphics[width=1.0\textwidth]{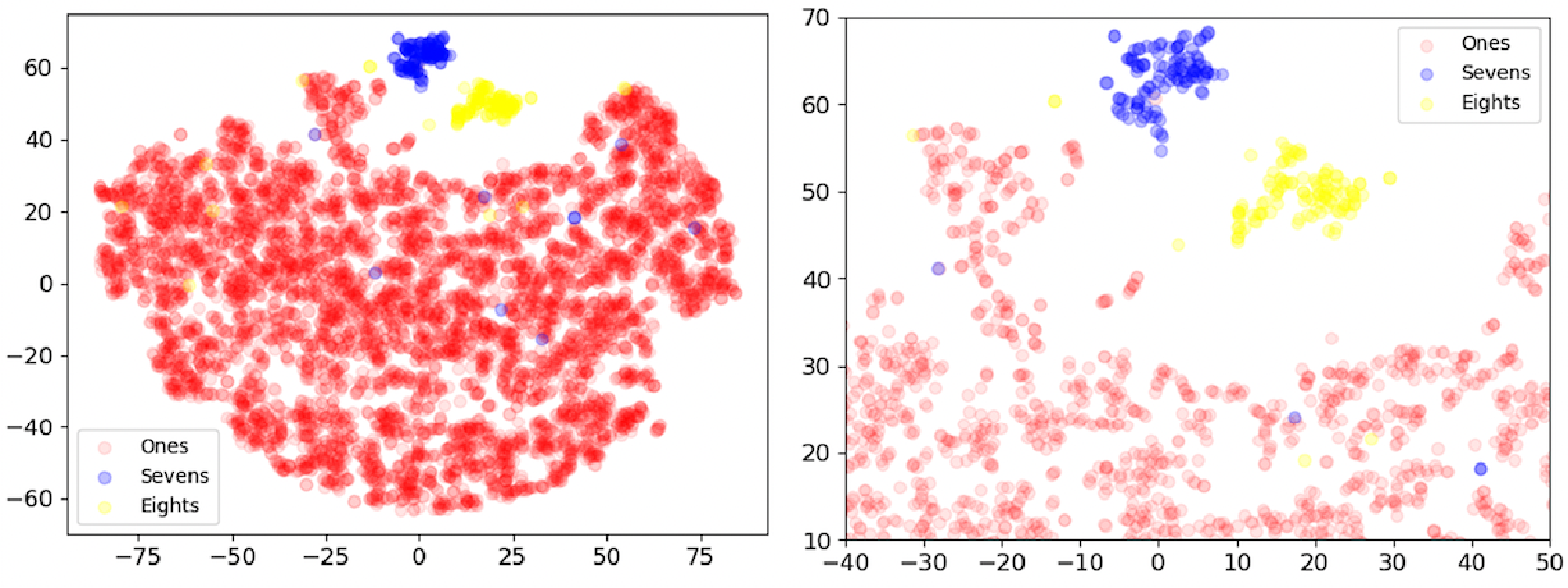}
    \caption{t-SNE 2D representation of distribution of dataset points upon the  PCA manifold $M=84$ with $1$s shown in red, $7$s in blue and $8$s in yellow.  We can see clearly the two “islands” where most of the $7$s and $8$s reside. This is shown zoomed in in the figure on the right. 
    }
    \label{fig:TSNE_PCA_all}
\end{figure*}

\begin{figure*}
    \centering
	\includegraphics[width=1.0\textwidth]{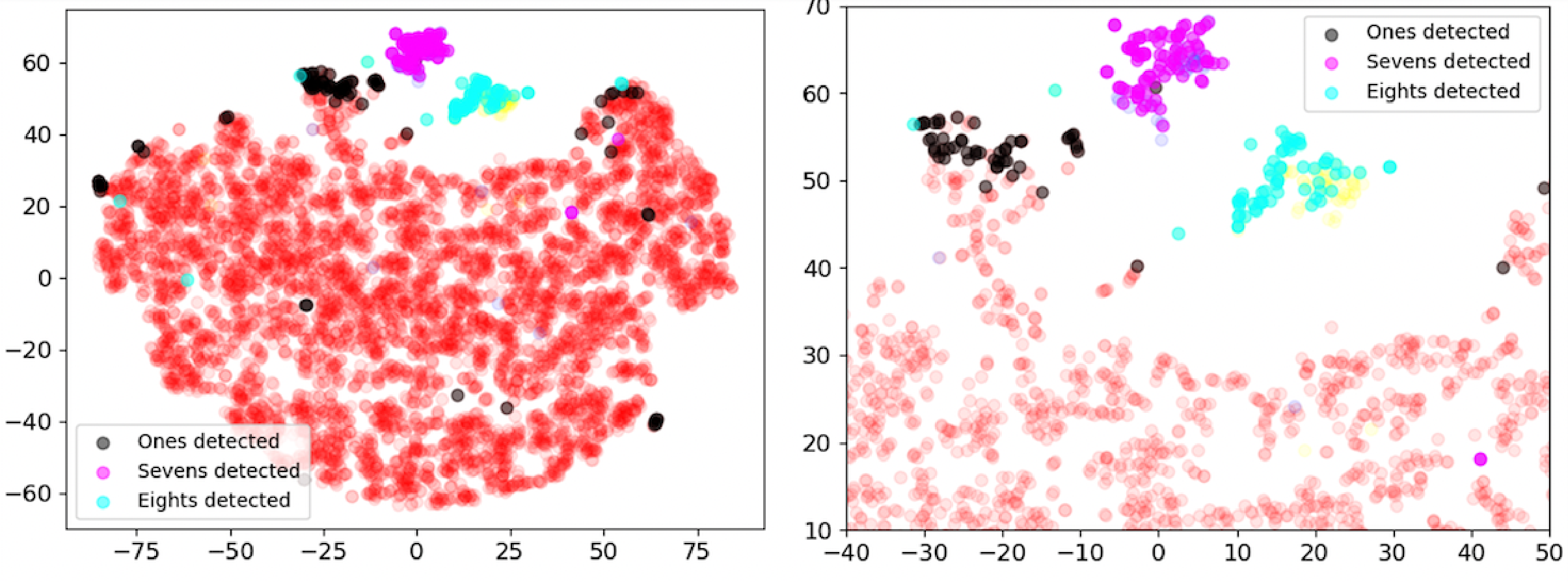}
    \caption{t-SNE 2D representation of distribution of dataset points upon the  PCA manifold $M=84$ with points detected as anomalies by off-manifold reconstruction error highlighted. Detected $7$s are shown in magenta, detected $8$s in cyan, and the detected $1$s (false positive anomalies) are shown in black. The subfigure to the right shows a zoomed-in view of this.
    We can see immediately that the reconstruction error approach is doing a decent job of identifying the “islands” of both $7$s and $8$s.}
    \label{fig:PCAdetanoms}
\end{figure*}

For each manifold, we then looked at the recall and precision of the unions of sets of anomalies from a) the off-manifold reconstruction error approach combined with each on-manifold approach in turn -- i.e. the Finding Pegasus Approach; and b) Isolation Forest combined with each of the other on-manifold detection methods in turn. This latter was done to see whether combining two on-manifold methods would give as good an outcome as our paper's proposed approach of combining complementary methods. Isolation Forest was chosen to pair with as it gave the best standalone results of the on-manifold methods for both AE and PCA manifolds. Results from manifold testing are shown in Tables~\ref{tab:ADtestPCA} and \ref{tab:ADtestAE}.

\begin{figure}
    \centering
	\includegraphics[width=0.99\columnwidth]{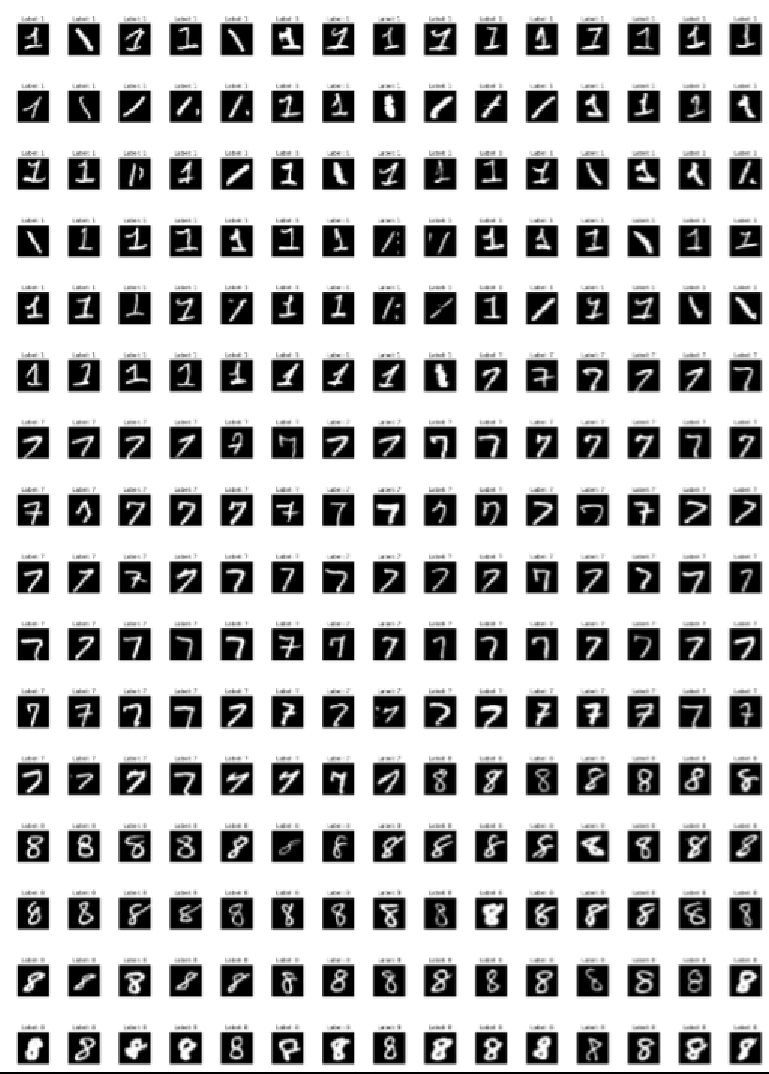}
    \caption{`Anomalies' detected from MNIST handwritten digits dataset using reconstruction error from PCA manifold at $M=84$   
    }
    \label{fig:PCA_RL_anomalies_digits}
\end{figure}

We also examined the composition of anomalies detected by on- and off-manifold methods across all four of the manifolds we constructed. e.g. the anomalies detected as the least well reconstructed points on the PCA manifold at $M=84$ is shown in 
Fig.~\ref{fig:PCA_RL_anomalies_digits}. Of the 240 worst-fitting points 89 are $7$s and 67 are $8$s. We note also how the $1$s which have been identified look different to a typical single vertical line.
The mix of $7$s and $8$s detected for each method with each manifold is given in Tables~\ref{tab:det78PCA} and ~\ref{tab:det78AE}.

Further details on manifold construction and visualisation of the data is given in the next section.

\begin{figure}
    \centering
	\includegraphics[width=0.99\columnwidth]{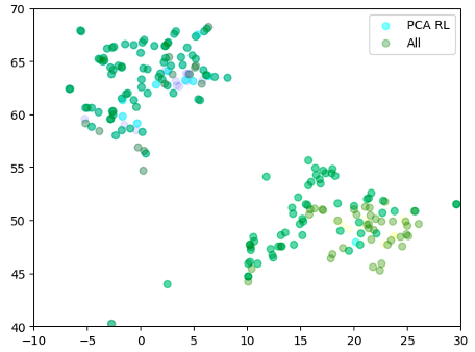}
    \caption{Zoomed-in view of TSNE 2D representation of distribution of dataset points upon the  PCA manifold $M=84$ showing those detected by PCA RE and those by all the on-manifold models combined. The points in cyan are anomalies that are only detected by the off-manifold method.}
    \label{fig:PCA_RL_Anomalies_v_therest}
\end{figure}

\subsection{Manifold Construction and Data Visualisation}\label{sec:maniconst}
\subsubsection{PCA Manifolds}

PCA manifolds were built using the Scikit-Learn implementation of PCA. The dimensionalities of the PCA manifolds were $M$=245 ($M\lesssim D$, c.f. $D$=784) with a data variance explained by the PCA decomposition of 99.5\%; and $M$=84 (putting it somewhere between the two regimes but closer to $M \ll D$) explaining 95\% of the data variance. This was deemd to be a reasonable level at which we could assume the manifold hypothesis still held.

\begin{figure*}
    \centering
	\includegraphics[width=1.0\textwidth]{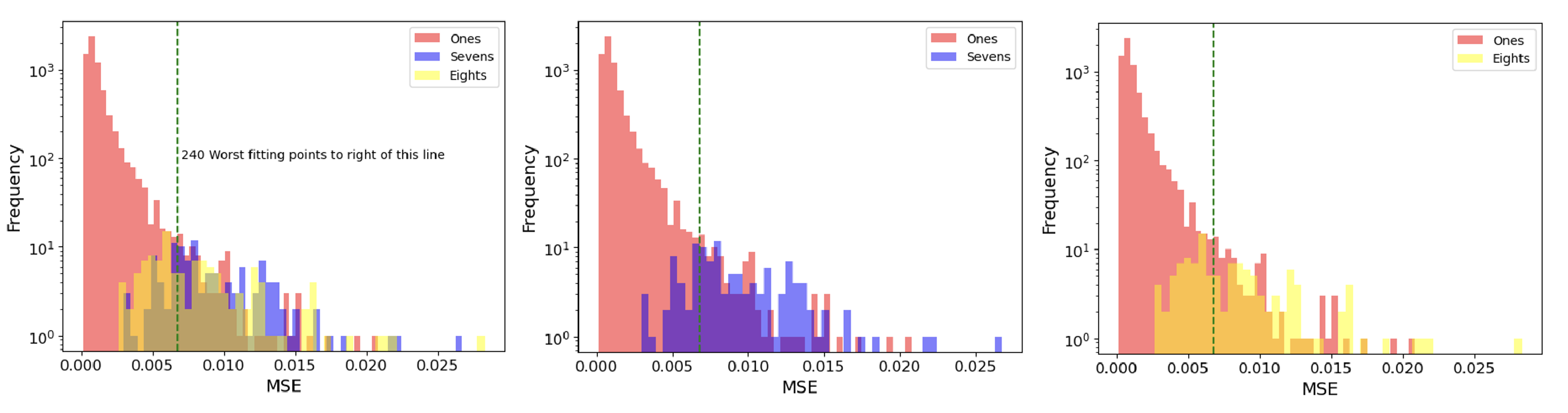}
    \caption{MSE of reconstruction error between PCA manifold and original data representations. Anomalies are defined as the most outlying 240 points} 
    \label{fig:PCA_allpoints}
\end{figure*}

We calculated the MSE between the original representation of each data point and its reconstruction from the PCA representation. Those which were worst represented had a higher MSE. The worst 240 points were assumed to be anomalies. The results for the $M=84$ manifold are shown in Fig.~\ref{fig:PCA_allpoints} with anomlaies to the right of the dashed green line.

If PCA reconstruction error were a perfect method of determining anomalies (these being defined as images of digits classified as $7$s and $8$s) then we would expect to see all of these to the right of the green line. The anomalies actually detected for the PCA manifold $M=84$ are shown in Fig.~\ref{fig:PCA_RL_anomalies_digits}. Of the 240 worst-fitting points 89 are $7$s and 67 are $8$s. We note also how the $1$s which have been identified look different to a typical single vertical line. 

We then built a 2D representation of the PCA manifold using t-SNE to see a representation of the distribution of the points upon it. For the $M=84$ manifold this is shown in Fig.~\ref{fig:TSNE_PCA_all} and  gives the  distribution of points representing $1$s (in red), $7$s (blue) and $8$s (yellow). 

In Fig~\ref{fig:PCAdetanoms} we plot the points we have detected as anomalies using the reconstruction error method.  We can see  that the reconstruction error approach performs reasonably well at identifying the “islands” of both $7$s and $8$s.

As discussed previously, we used various unsupervised AD methods -- LOF, EllipticEnvelope, One-Class SVM, and Isolation Forest -- on the PCA generated manifold to find outlying points; and then combined them with IF and PCA Reconstruction to test AD recall, and these results are shown in Table~\ref{tab:ADtestPCA}.

We also combined PCA $M=84$ reconstruction error with a combination of IF and LOF and this achieved a recall of 85\% (cf. 78.8\% for IF-LOF) and precision 44.9\% (cf 44.7\% for IF-LOF), showing again the value of combining on and off-manifold techniques.

Even when \textit{all} the on-manifold methods were combined for the PCA $M=84$ manifold, there were still some anomalies which are detected by the off-manifold method which the combination of on-manifold methods do not detect (the points shown in cyan in Fig.~\ref{fig:PCA_RL_Anomalies_v_therest}).

\begin{figure}
    \centering
	\includegraphics[width=0.99\columnwidth]{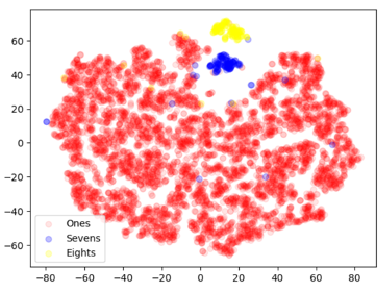}
    \caption{Chart showing distribution of different underlying data classes on t-SNE 2D representation of the AE $M=30$ manifold}
    \label{fig:TSNE_AE_data}
\end{figure}

\begin{figure}
    \centering
	\includegraphics[width=0.99\columnwidth]{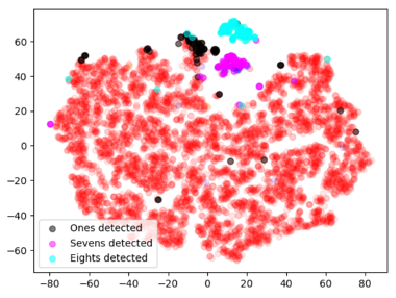}
    \caption{Chart showing distribution of anomalies detected through AE reconstruction error from the AE $M=30$ manifold}
    \label{fig:TSNE_AE_anomalies}
\end{figure}

Distribution of the anomalies detected on the t-SNE representation are shown in Fig~\ref{fig:TSNE_AE_anomalies}.

\subsubsection{AE Manifolds}
This work was then repeated using AEs.
AE manifolds were built using Keras (Tensorflow) \cite{chollet2015keras} with one hidden layer in each of the encoder and decoder. The dimensionalities of the AE manifolds were $M$=245 ($M\lesssim D$, c.f. $D$=784)  and $M$=30. The latter was following Geron \yrcite{geron2023} enabling us to use only 30 dimensions for the latent space  and still satisfy the manifold hypothesis.

\begin{figure}
    \centering
	\includegraphics[width=0.99\columnwidth]{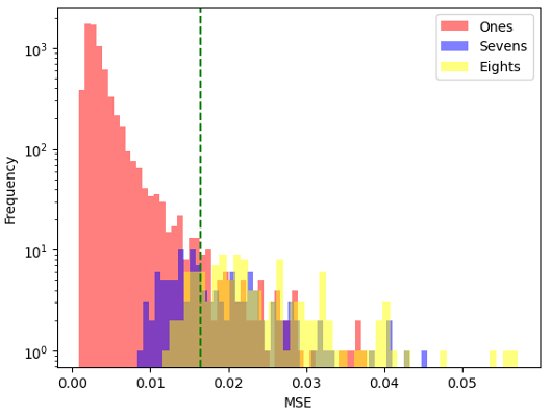}
    \caption{Chart showing MSE between reconstruction and original data from AE generated manifold}
    \label{fig:AE_anomalies}
\end{figure}

Using reconstruction error to identify anomalies, the distribution of MSE for the dataset is shown in  Fig.~\ref{fig:AE_anomalies}.

The distribution of the $1$s, $7$s and $8$s on a t-SNE view of the $M=30$ AE manifold is shown in Fig.~\ref{fig:TSNE_AE_data}. Note how the 2D representation of the manifold is not identical to that of the PCA manifold shown in Fig.~\ref{fig:TSNE_PCA_all}, demonstrating how the manifold construction is model dependent.

%Even better definition of the anomalies suggesting that the non-linear nature of the data is handled well by a manifold learning method.

We then employed the same methods that were used with PCA to perform AD on the AE manifolds. We computed recall and precision for these techniques standalone and in conjunction with AE reconstruction error based approach, as previously discussed and as shown in Table~\ref{tab:ADtestAE}.

%%%%%%%%%%%%%%%%%%%%%%%%%%%%%%%%%%%%%%%%%%%%%%

%%%%%%%%%%%%%%%%%%%%%%%%%%%%%%%%%%%%%%%%%%%%%%%%%%

%%%%%%%%%%%%%%%%%%%%%%%%%%%%%%%%%%%%%%%%%%%%%%%%%%%%%%%%%%%%%%%%%%%%%%%%%%%%%%%

\end{document}